\documentclass[sigconf]{aamas}

\usepackage{balance} % for balancing columns on the final page
\usepackage{svg} % for including SVG images
\usepackage{subcaption} % for subfigure environment
\usepackage{xcolor}
\usepackage{multirow}
\usepackage[section]{placeins}

\doi{KXRD5206}

\makeatletter
\gdef\@copyrightpermission{
	\begin{minipage}{0.2\columnwidth}
		\href{https://creativecommons.org/licenses/by/4.0/}{\includegraphics[width=0.90\textwidth]{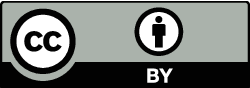}}
	\end{minipage}\hfill
	\begin{minipage}{0.8\columnwidth}
		\href{https://creativecommons.org/licenses/by/4.0/}{This work is licensed under a Creative Commons Attribution International 4.0 License.}
	\end{minipage}
	\vspace{5pt}
}
\makeatother

\setcopyright{ifaamas}
\acmConference[AAMAS '26]{Proc.\@ of the 25th International Conference
	on Autonomous Agents and Multiagent Systems (AAMAS 2026)}{May 25 -- 29, 2026}
{Paphos, Cyprus}{C.~Amato, L.~Dennis, V.~Mascardi, J.~Thangarajah (eds.)}
\copyrightyear{2026}
\acmYear{2026}
\acmDOI{}
\acmPrice{}
\acmISBN{}

\title[AAMAS-2026 Formatting Instructions]{AltNet: 
Addressing the Plasticity-Stability Dilemma in Reinforcement Learning}

\author{Mansi Maheshwari}
\authornote{Corresponding author.}
\affiliation{
  \institution{University of Massachusetts}
  \city{Amherst}
  \country{United States}}
\email{mmaheshwari@umass.edu}

\author{John C. Raisbeck}
\affiliation{
  \institution{University of Massachusetts}
  \city{Amherst}
  \country{United States}}
\email{jraisbeck@umass.edu}

\author{Bruno Castro da Silva}
\affiliation{
  \institution{University of Massachusetts}
  \city{Amherst}
  \country{United States}}
\email{bsilva@umass.edu}

\begin{abstract}
Artificial neural networks have shown remarkable success in supervised learning when trained on a single task using a fixed dataset. However, when neural networks are trained on a reinforcement learning task, their ability to continue learning from new experiences declines over time. This decline in learning ability is known as plasticity loss. To restore plasticity, prior work has explored periodically resetting the parameters of the learning network, a strategy that often improves performance. However, such resets come at the cost of a temporary drop in performance, which can be dangerous in real-world settings. To overcome this instability, we introduce AltNet, a reset-based approach that restores plasticity without performance degradation by leveraging a pair of ``twin'' networks. The use of twin networks anchors performance during resets through a mechanism that allows networks to periodically alternate roles: one network learns as it acts in the environment, while the other learns off-policy from the active network’s interactions through a replay buffer. At fixed intervals, the active network is reset and the passive network, having learned from prior experience, becomes the new active network. AltNet restores plasticity, improving sample efficiency and achieving higher performance, while avoiding performance drops that pose risks in safety-critical settings. We demonstrate these advantages in several high-dimensional control tasks from the DeepMind Control Suite, where AltNet outperforms various relevant baseline methods, as well as state-of-the-art reset-based techniques.
\end{abstract}

\keywords{Plasticity; Stability; Plasticity-Stability Dilemma; Continual Learning; Reinforcement Learning; Network Resets}

\newcommand{\extraParagraphSpacing}{}%{\\}
         
\newcommand{\BibTeX}{\rm B\kern-.05em{\sc i\kern-.025em b}\kern-.08em\TeX}

\acmBadgeL{}
\acmBadgeR{}

\begin{document}

%%% The following commands remove the headers in your paper. For final 
%%% papers, these will be inserted during the pagination process.

\pagestyle{fancy}
\fancyhead{}

%%% The next command prints the information defined in the preamble.

\maketitle 

%%%%%%%%%%%%%%%%%%%%%%%%%%%%%%%%%%%%%%%%%%%%%%%%%%%%%%%%%%%%%%%%%%%%%%%%

\section{Introduction}

Many deep learning systems are designed to be trained on a single task and to converge to a single solution. In non-stationary environments, however, the goal being optimized by the model evolves over time. Success in such settings requires continual adaptation rather than the ability to identify a single solution. This need motivates the fields of continual and lifelong learning, where an agent updates, accumulates, and exploits knowledge throughout its lifetime \citep{chen2018lifelong}. A central obstacle in continual learning is \emph{plasticity loss}---the progressive decline in an agent’s ability to learn from new data over time \citep{nikishin2022primacy, lyle2022understanding, dohare2024loss, kumar2020implicit}. We say that a network has lost plasticity if its performance can no longer be improved as effectively as a freshly initialized counterpart~\citep{lyle2024disentangling}. Plasticity loss has been observed in non-stationary settings. For instance, \citet{achille2017critical} showed that pre-training on blurred CIFAR images impaired subsequent learning of the original dataset. Similarly, \citet{ash2020warm} found that pre-training on half of a dataset and using the resulting model as a starting point when tackling a supervised learning task reduced accuracy compared to training on the full dataset from scratch. More broadly, \citet{dohare2021continual} demonstrated that when neural networks are trained sequentially on multiple tasks, their ability to learn new tasks declines with each additional task.\extraParagraphSpacing{}

Reinforcement learning (RL) compounds the difficulty of maintaining plasticity over time because, even when the task itself is stationary, RL agents face inherent sources of non-stationarity. First, in online RL, agents collect their own data; as policies evolve, the distribution of encountered states and actions shifts, producing \emph{input non-stationarity}. Second, many RL algorithms such as DQN, A2C, PPO, and SAC \citep{mnih2015human, mnih2016asynchronous, schulman2017proximal, haarnoja2018soft} rely on bootstrapping, where predictions of future rewards themselves serve as learning targets. As these predictions evolve, the targets themselves change, creating \emph{target non-stationarity}. Together, these factors require agents to continually adapt to shifting data distributions even when tackling a single task, thereby amplifying plasticity loss.\extraParagraphSpacing{}

To mitigate plasticity loss, various approaches have been proposed (\autoref{sec:related}). Among these, a particularly promising family of methods is based on periodically resetting network parameters \citep{nikishin2022primacy, kim2023sample, dohare2024loss, sokar2023dormant}. Resets are effective because they restore the network to a well-conditioned, highly plastic initialization that is gradually lost during training. As networks adapt to specific tasks or data distributions, they accumulate pathologies---such as dormant neurons, increasing weight magnitudes, and reduced rank---that impair their ability to learn from new data \citep{dohare2024loss}. Resetting the parameters removes these accumulated effects and reinitializes the network to conditions resembling its original, plastic initialization. 
%(see supporting analysis in~\hyperref[sec:appendix_plasticity]{\autoref{sec:appendix_plasticity}}). 
\citet{nikishin2022primacy} empirically demonstrated that resetting a network can substantially improve performance by renewing its ability to learn and exploit data. Although effective, full network resets come at a cost: they discard learned models and can cause an immediate performance drop (for an example, see \autoref{fig:rr1}, orange curve). This makes Standard Resets \citep{nikishin2022primacy} impractical for real-world deployment. In this paper, we address the challenge of retaining the benefits of full network resets in restoring plasticity while avoiding the performance instability they induce.\extraParagraphSpacing{}

To address the plasticity-stability dilemma, we introduce \emph{AltNet}, a reset-based alternating network approach that preserves plasticity without inducing recurring performance drops. AltNet maintains two networks that periodically switch roles. At any given time, the \emph{active network} interacts with the environment, while the \emph{passive network} learns off-policy from the active agent’s experience and a shared replay buffer. At fixed intervals, the active network is reset and the passive network, having learned from prior experience, becomes the new active network. This alternating structure anchors performance during resets and prevents performance collapse. Importantly, AltNet successfully leverages resets while significantly reducing performance instability even with little experience replay; in these cases, by contrast, Standard Resets \citep{nikishin2022primacy} fail and more sophisticated methods such as Resets with Deep Ensembles (RDE) \citep{kim2023sample} still exhibit sharp post-reset performance drops (see \autoref{fig:rr1}). To understand which factors contribute to AltNet's superior performance, we systematically evaluate aspects such as model capacity, number of networks, replay ratio, buffer size, and reset duration (\autoref{sec:abl_buffer}). Finally, we show that AltNet also improves performance in on-policy settings, as demonstrated by comparisons with relevant on-policy baselines (\autoref{sec:ppo}).

%%%%%%%%%%%%%%%%%%%%%%%%%%%%%%%%%%%%%%%%%%%%%%%%%%%%%%%%%%%%%%%%%%%%%%%%

\section{Related Work}
\label{sec:related}
\textbf{Plasticity. } Prior work uses the term \emph{plasticity} to refer to the degree to which a network generalizes to unseen data \cite{berariu2021study} or to refer to its ability to continue improving performance on its training objective over time \citep{abbas2023loss, nikishin2023deep, kumar2020implicit, lyle2024disentangling}. In this paper, we adopt the latter meaning.  We say that a network has lost plasticity if its performance cannot be improved as effectively as a freshly initialized counterpart.\\

\noindent\textbf{Plasticity loss in reinforcement learning. } Several prior works point to the same underlying challenge: neural networks in reinforcement learning often lose their ability to adapt as training progresses. \citet{lyle2022understanding} observe a gradual loss of capacity to fit evolving targets even in a single task, while \citet{kumar2020implicit} attribute a similar effect to implicit under-parameterization. \citet{nikishin2022primacy} introduce the term \emph{primacy bias}, referring to the tendency of agents to overfit to early experiences, which hinders subsequent learning. Although framed in different ways, these studies describe facets of the same phenomenon---plasticity loss. We choose the terminology \emph{plasticity loss} because it captures the common thread: the gradual decline in an agent’s ability to adapt to new information.\\\extraParagraphSpacing{}

\noindent\textbf{Causes of plasticity loss. } The precise cause of plasticity loss remains unknown. Several correlates have been identified, such as inactive neurons, growth of the network's average weight magnitude, decrease in the expressivity of the network, and changes in the curvature of the loss landscape \citep{sokar2023dormant, dohare2021continual, kumar2020implicit, lyle2023understanding}. None of the proposed correlates, however, provide a consistent explanation across settings. For example, \citet{lyle2023understanding} show that for any proposed correlate, counterexamples can be constructed where the correlation disappears or even reverses. Since the underlying cause of plasticity loss has not been identified, it is difficult to directly determine whether a system has retained or lost plasticity. Following prior work \citep{nikishin2022primacy, nikishin2023deep, kim2023sample}, we use performance as a proxy for plasticity.\extraParagraphSpacing{}

A wide range of strategies have been proposed to mitigate plasticity loss. Broadly, these fall into two families: methods based on regularization, which constrain or perturb weights to preserve plasticity, and methods based on resets, which periodically reinitialize parts or the entire network. Below, we briefly review each in turn:
\begin{enumerate}
    \item \textbf{Regularization-based strategies. } Prior work has explored regularization-based strategies to maintain plasticity. While $\ell^{1}$ and $\ell^{2}$ weight penalties can slow down weight growth, they sometimes aggravate rank collapse by biasing weights toward the origin \citep{dohare2021continual, lyle2023understanding}. To address this, methods such as Shrink-and-Perturb \citep{ash2020warm} and L2 Init \citep{kumar2020implicit} have been proposed, which encourage weight updates toward high-plasticity initializations while preserving feature diversity.\extraParagraphSpacing{}

    \item \textbf{Reset-based strategies. }
    Another family of methods directly resets the network in whole or in part. Continual Backprop \citep{dohare2024loss} and ReDO \citep{sokar2023dormant} reset subsets of neurons selected for low utility or persistent inactivity. \citet{igl2020transient} propose distilling a trained policy into a newly initialized network, which can be seen as a form of reset with distillation as the transfer mechanism. \citet{nikishin2022primacy} propose periodic full network resets, relying on a replay buffer to transfer knowledge, but these cause sharp performance drops. In this paper, we refer to this approach as Standard Resets.
    Reset Deep Ensembles (RDE) \citep{kim2023sample} leverages full network resets by maintaining an ensemble of networks, resetting one network at a time to induce plasticity. Actions are chosen through a $Q$-value–weighted voting scheme, where each proposed action is weighted by the critic of the oldest network in the ensemble. Although RDE improves stability over Standard Resets \citep{nikishin2022primacy}, it still suffers from significant post-reset performance drops because a freshly reset, untrained network can still act in the environment.
\end{enumerate}
%%%%%%%%%%%%%%%%%%%%%%%%%%%%%%%%%%%%%%%%%%%%%%%%%%%%%%%%%%%%%%%%%%%%%%%%
\section{AltNet}
\label{sec:method}
In this section, we introduce AltNet, a dual-network reset-based architecture designed to restore plasticity while maintaining stability in reinforcement learning.\\\extraParagraphSpacing{}

\noindent \textbf{Central Hypothesis. } 
Prior work has shown that resets can restore plasticity \citep{nikishin2022primacy}, but they also cause sharp performance collapses when the reset policy acts immediately. We hypothesize that leveraging two insights can reconcile this plasticity-stability dilemma: (i) resetting a network initializes it to a highly plastic state, from which it may be possible to learn a better policy, as compared to a trained network, and  (ii) using well-trained networks for interaction with the environment prevents performance drops. To combine the benefits of both, AltNet introduces a dual-network architecture that allows resets to occur while avoiding performance instability.\\ \extraParagraphSpacing{}

\begin{figure}[t]
    \centering
    \includegraphics[width=\linewidth]{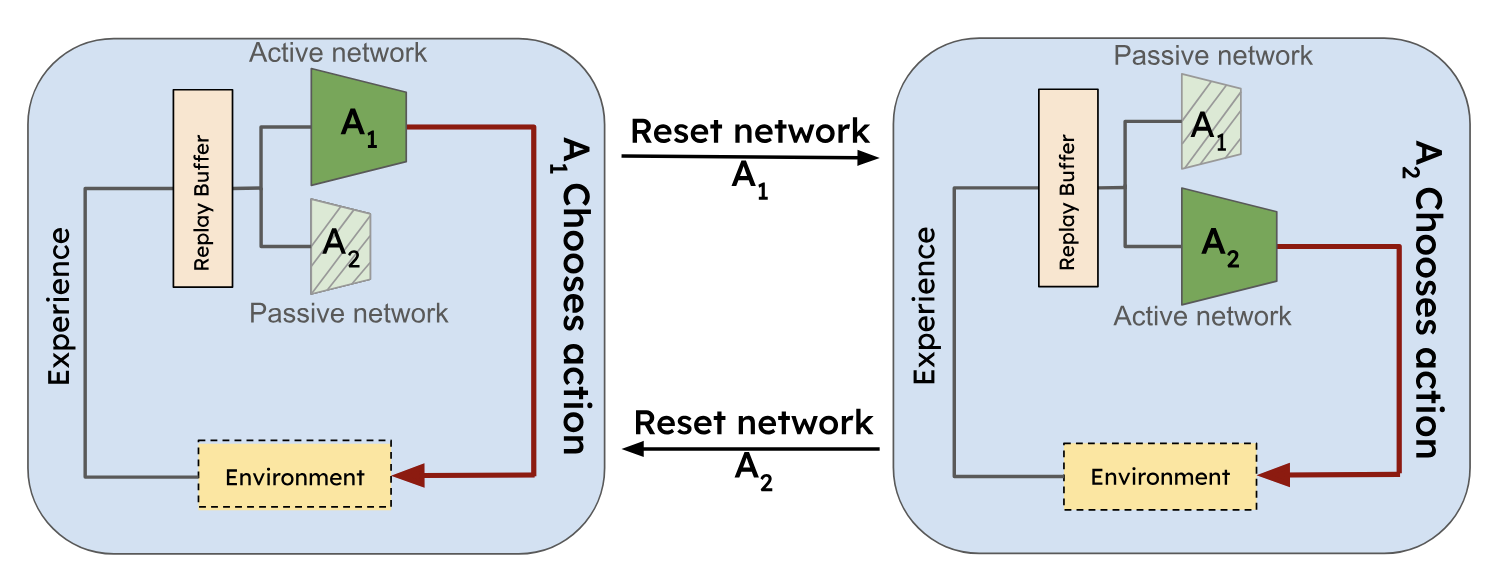}
    \caption{AltNet maintains two networks, A$_{1}$ and A$_{2}$, which share a replay buffer and alternate roles over time. Initially, A$_{1}$ (dark green) is active and collects experience by directly interacting with the environment, while A$_{2}$ (light green) remains passive and undergoes off-policy updates. At every \texttt{ResetFreq} steps, the active network is reset and becomes passive, while the previously passive network becomes active. This cyclic alternation enables frequent resets to maintain plasticity without sacrificing stability.}
    \label{fig:altNet}
\end{figure}

\noindent \textbf{Architecture. } 
AltNet is composed of two networks that alternate roles at a fixed interval, \texttt{ResetFreq} (\autoref{fig:altNet}). At any given time, the active network interacts with the environment, while the passive network learns off-policy from the experiences of the active network and a replay buffer. The replay buffer is shared between the twin networks. Every \texttt{ResetFreq} steps the active network is reset and becomes the passive network and vice versa. This alternating cycle ensures that resets occur frequently enough to counter plasticity loss, yet performance remains stable because only trained networks interact with the environment.\\\extraParagraphSpacing{}

\noindent \textbf{Key Innovation. }
AltNet makes a structural departure from prior reset-based approaches. It prevents recently-reset networks from acting in the environment until they have received sufficient training. In contrast, Standard Resets \citep{nikishin2022primacy} expose the reset network directly to the environment, making immediate performance collapse inevitable. RDE \citep{kim2023sample} employs ensembles with a Q-value–weighted gating policy to reduce the likelihood that a reset agent acts prematurely, but still allows recently reset networks to act. AltNet, on the other hand, guarantees that only trained networks interact with the environment. In AltNet, reset networks first train passively before taking over. Empirically, the resulting mechanism is more robust and simpler: AltNet avoids post-reset performance drops across replay ratios and achieves higher and more stable returns (see, e.g., \autoref{fig:rr1}). More broadly, AltNet shows that even with full network resets, plasticity and stability can be simultaneously achieved. 
%%%%%%%%%%%%%%%%%%%%%%%%%%%%%%%%%%%%%%%%%%%%%%%%%%%%%%%%%%%%%%%%%%%%%%%%
\section{Results and Analysis}
\label{sec:results}

In this section, we present empirical evidence that AltNet effectively addresses the plasticity–stability dilemma. We begin by contextualizing its performance through comparisons with established reset-based and non-reset baselines in diverse continuous-control environments. Next, we analyze the contributions of replay buffer preservation, network alternation, and periodic resets to higher performance. Finally, we extend our investigation to on-policy settings, where replay buffers are absent, to examine how alternating resets influence learning dynamics in these regimes.\extraParagraphSpacing{}

%\vspace{-3ex}
\begin{figure}[t]
    \centering    
    \includegraphics[width=1\linewidth]{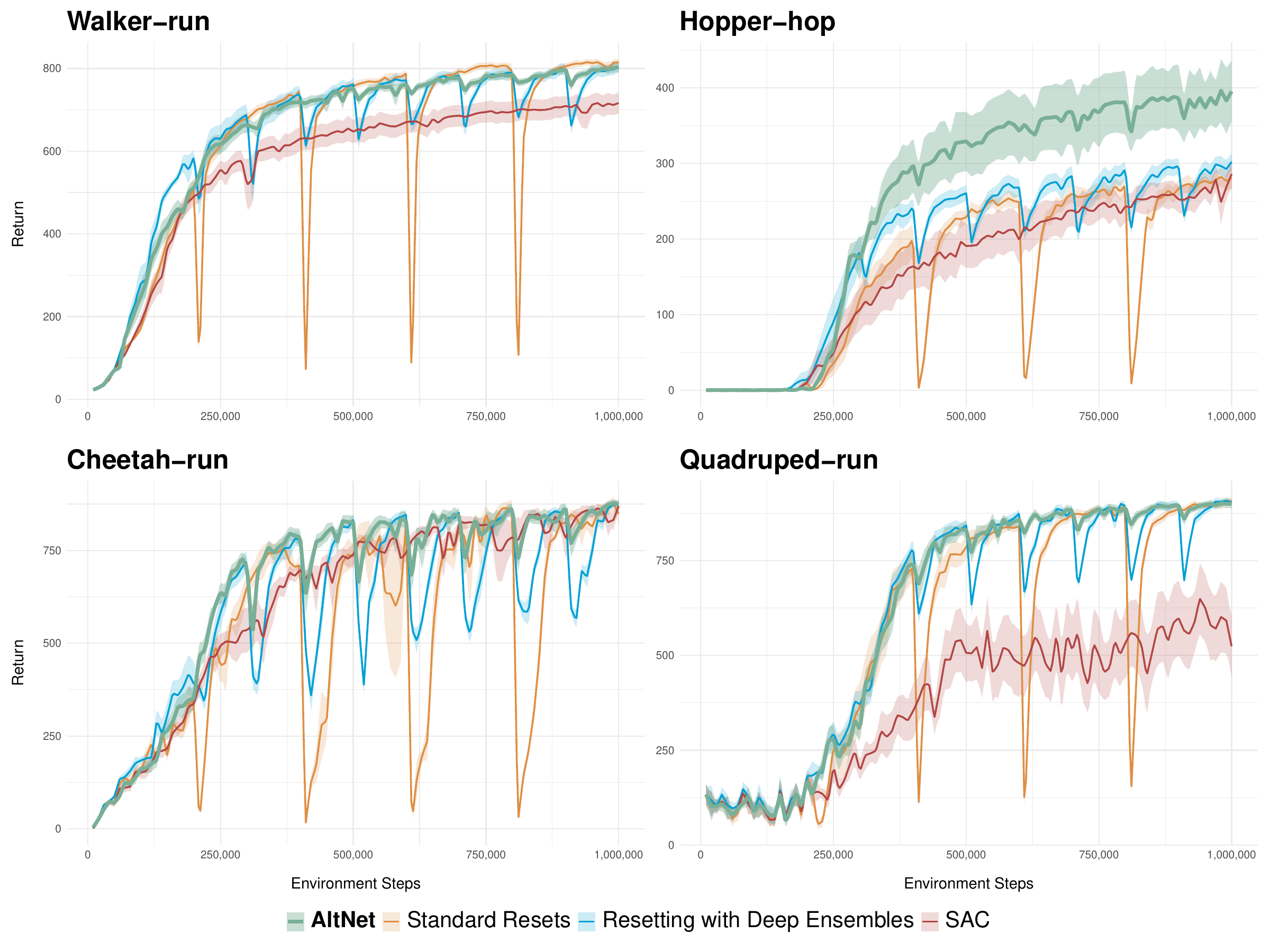} 
    \caption{Learning curves in four DMC environments when replay ratio = 1. Results are averaged over 10 seeds; shaded regions indicate $\pm$1 standard error. AltNet (green) avoids post-reset drops and achieves higher returns compared to SAC (red), Standard Resets (orange), and RDE (blue).}
    \label{fig:rr1}
    %\vspace{-2ex}
\end{figure}

We first evaluate AltNet on stationary continuous-control tasks from the DeepMind Control Suite \citep{tassa2018deepmind}. We focus on the off-policy setting and use Soft Actor–Critic (SAC) \citep{haarnoja2018soft} as the underlying algorithm for all methods, and as our baseline. SAC leverages a replay buffer to reuse past experiences. Later, we combine AltNet with PPO to assess whether our method's benefits extend to on-policy settings, which do not rely on a replay buffer.\extraParagraphSpacing{}

All agents are trained for 1M environment interactions. The reset interval is fixed at \(200{,}000\) gradient updates (\(U\)). We vary the replay ratio \(\text{RR} \in \{1,4\}\), defined as the number of gradient updates per environment step. To ensure a fair comparison, we adopt the reset frequency normalization proposed by \citet{kim2023sample}, which accounts for both the replay ratio ($\text{RR}$) and the number of networks ($N$), as shown in Eq.~\eqref{eq:resetfreq}. We compare AltNet against the non-reset-based baseline SAC~\citep{haarnoja2018soft}, as well as state-of-the-art full network reset-based methods: Standard Resets~\citep{nikishin2022primacy} and Resets with Deep Ensembles (RDE)~\citep{kim2023sample}.
\begin{equation}
    \texttt{ResetFreq}_{\text{(env steps)}} = \frac{U}{\text{RR} \times N}
    \label{eq:resetfreq}
\end{equation}
As shown in \autoref{fig:rr1}, when $\text{RR} = 1$, Standard Resets' performance collapses almost immediately, while RDE exhibits sharp post-reset drops in average return. In contrast, AltNet avoids these instabilities by anchoring performance during resets through a twin network that assumes control after each reset. Consequently, AltNet achieves higher average returns and smoother learning curves across tasks.\extraParagraphSpacing{}

\begin{figure}[ht!!]
    \centering    
    \includegraphics[width=1\linewidth]{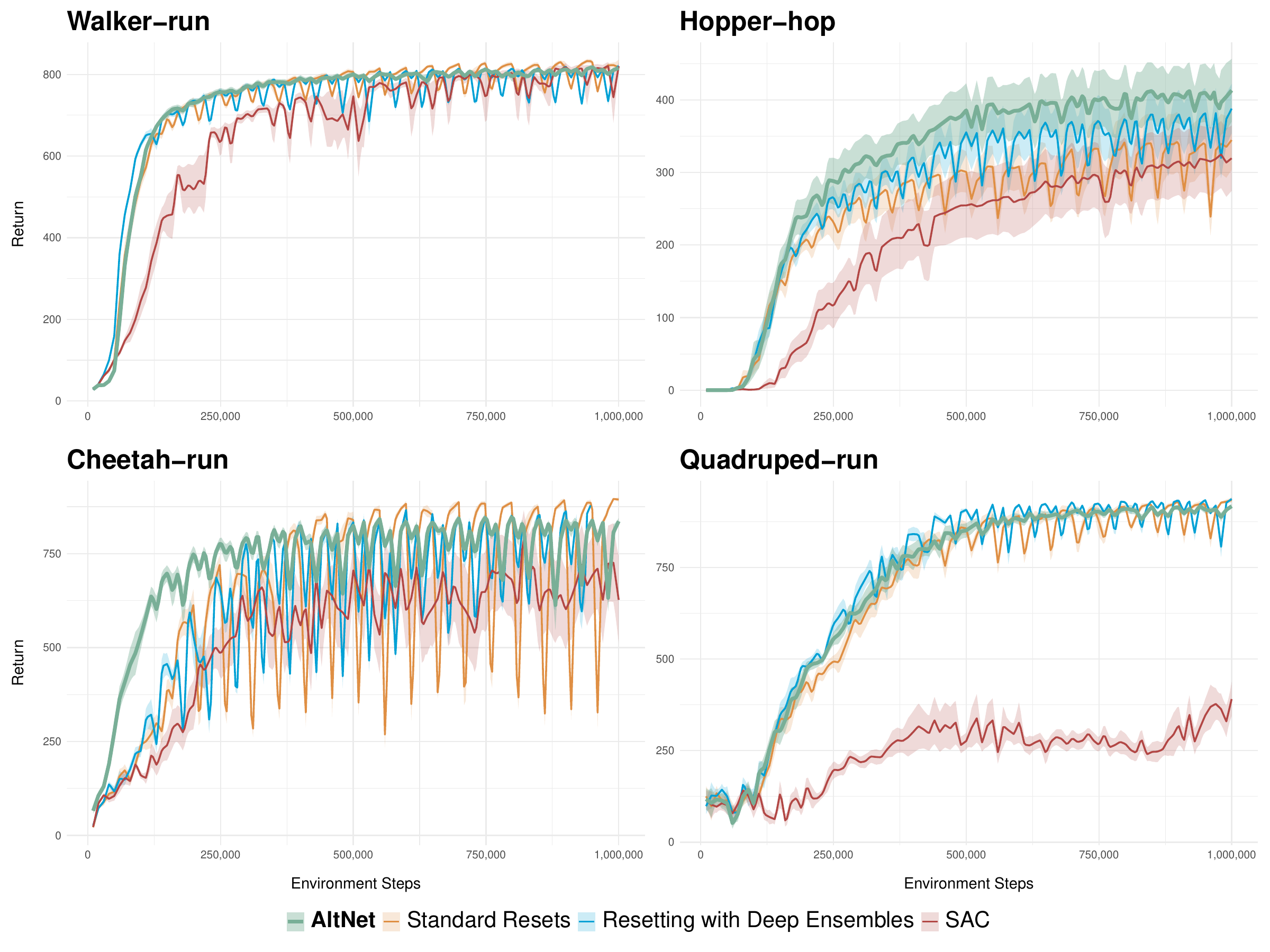}
    \caption{Learning curves in four DMC environments when replay ratio = 4. Results are averaged over 10 seeds; shaded regions indicate $\pm$1 standard error. AltNet (green) avoids post-reset drops and achieves higher returns compared to SAC (red), Standard Resets (orange), and RDE (blue).}
    \label{fig:rr4}
    %\vspace{-3ex}
\end{figure}

% Point 8
When $\text{RR} = 4$, AltNet continues to outperform SAC and Standard Resets while maintaining stable training dynamics and avoiding post-reset performance drops. In all environments except one, AltNet matches or exceeds the performance of RDE (see \autoref{tab:auc_all}). In Quadruped-run, RDE achieves a slightly higher normalized AUC, while AltNet yields more stable training dynamics (\autoref{fig:rr4}).\extraParagraphSpacing{}

% GRID_2X2_RR_4
% M_RR4_c1find
%\vspace{-5ex}

We also report, in \autoref{tab:auc_all}, the mean normalized \emph{area under the learning curve} (AUC), which summarizes the overall learning trajectory of an agent. The AUC integrates performance over time, capturing both higher returns and greater stability: agents that maintain strong performance throughout training achieve a higher AUC. We normalize the AUC to rescale the values into a consistent range, making comparisons more interpretable. See ~\autoref{sec:auc} for details on the AUC calculation. As shown in \autoref{tab:auc_all}, \mbox{AltNet} achieves the highest AUC in 7 out of 8 combinations of environments and replay ratios, outperforming SAC by approximately $38\%$, SR by $12\%$, and RDE by $6\%$ on average.\extraParagraphSpacing{}

% \begin{table} [t]
\begin{table}[t]
\centering
\caption{Normalized average AUC of different methods across DMC environments. The best method in each environment is highlighted in bold. AltNet achieves the highest normalized AUC, outperforming SAC by approximately $38\%$, SR by $12\%$, and RDE by $6\%$ on average.}
\begin{tabular}{lcccc}
\toprule
\textbf{Environment} & \textbf{AltNet} & \textbf{RDE} & \textbf{SAC} & \textbf{SR} \\
\midrule
Cheetah (RR=1)   & \textbf{658.27} & 596.62 & 616.12 & 529.94 \\
Hopper (RR=1)    & \textbf{248.68} & 189.79 & 156.69 & 154.52 \\
Quadruped (RR=1) & \textbf{619.12} & 609.36 & 377.27 & 568.36 \\
Walker (RR=1)    & \textbf{645.76} & 643.22 & 570.08 & 617.06 \\
\midrule
Cheetah (RR=4)   & \textbf{721.85} & 619.15 & 535.80 & 595.13 \\
Hopper (RR=4)    & \textbf{313.78} & 278.29 & 205.00 & 248.94 \\
Quadruped (RR=4) & 703.74 & \textbf{717.24} & 240.93 & 687.43 \\
Walker (RR=4)    & \textbf{728.49} & 723.64 & 653.82 & 725.44 \\
\midrule
Average (RR=1) & \textbf{542.96} & 509.75 & 430.04 & 467.47 \\
Average (RR=4) & \textbf{616.97} & 584.61 & 408.89 & 565.24 \\
\bottomrule
\end{tabular}
\label{tab:auc_all}
\end{table}

\subsection{AltNet Enhances Sample Efficiency}
\label{sec:sample_efficiency}
In this section, we evaluate whether AltNet improves performance and sample efficiency compared to an SAC baseline. We also compare AltNet against SAC trained at higher replay ratios---a common strategy to improve sample efficiency.\extraParagraphSpacing{}

\begin{figure}[hb!!]
    \centering
    \includegraphics[width=1\linewidth]{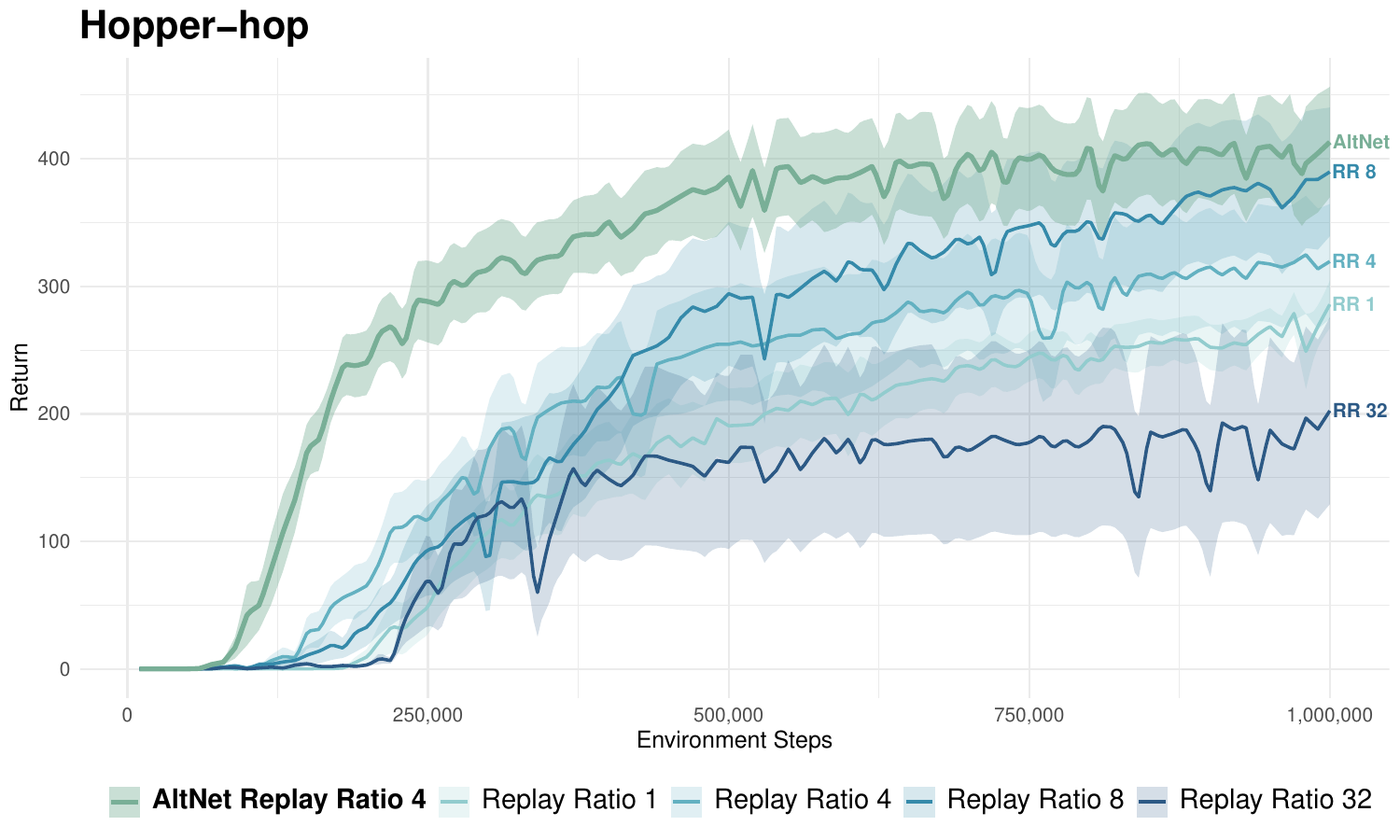}
    \caption{Learning curves of SAC in the \texttt{hopper-hop} environment (DMC) under different replay ratios (RR = 1, 4, 8, 32) and AltNet at RR = 4. Curves show mean episodic return over 10 seeds, with shaded regions denoting $\pm$1 standard error. For SAC, performance improves as RR increases up to 8, but degrades at RR = 32. AltNet achieves the highest performance at RR = 4 and is the most sample efficient.}
    \label{fig:highRR}
    %\vspace{-1ex}
\end{figure}

In reinforcement learning, agents learn through direct interaction with the environment, which is often slow and expensive in real-world domains such as robotics or healthcare applications. This makes \emph{sample efficiency}---learning as much as possible from limited interactions---a central concern. A common strategy to improve sample efficiency is to increase the \emph{replay ratio} (RR), defined as the number of gradient updates performed per environment step \citep{fedus2020revisiting, wang2016sample, d2022sample}. Higher replay ratios allow agents to reuse past experiences more extensively, thereby extracting additional learning signals from limited data. However, increasing RR also linearly increases computational cost and, beyond a point, can degrade performance due to overfitting to outdated experiences \citep{nikishin2022primacy, d2022sample}. As shown in \autoref{fig:highRR}, increasing the replay ratio from 1 to 8 improves SAC’s performance by allowing more updates per sample, but performance is substantially lower when $\text{RR}=32$. AltNet, by contrast, achieves superior performance even at $\text{RR}=1$ (see \autoref{fig:all_curves}) and $\text{RR}=4$ (see \autoref{fig:highRR}), surpassing SAC trained at much higher replay ratios. This shows that unlike SAC, AltNet achieves higher performance and greater sample efficiency at lower replay ratios, and consequently reduced computational overhead. We further establish that AltNet's improved performance is not due to increased capacity from the additional network: reducing the total number of parameters across the two networks to match a single SAC network yields nearly identical performance (\autoref{fig:vari}).\extraParagraphSpacing{}

To quantify sample efficiency directly, we measure the \emph{average return at fixed interaction budgets} of 100k, 300k, and 500k steps.
% and the \emph{time required to reach a fixed performance threshold}. 
%
%AltNet consistently achieves higher returns at every fixed budget and reaches the target threshold substantially earlier than SAC across replay ratios (\autoref{tab:fixed_budget}). 
%
AltNet consistently (and more rapidly, i.e., for all interaction budgets) achieves higher returns than SAC across all evaluated replay ratios. 
% ×
In particular, it outperforms the best SAC baselines by \textbf{52}x at 100k interactions, \textbf{1.8}x at 300k, and \textbf{1.3}x at 500k ~(\autoref{tab:fixed_budget}).
%
%In particular, it outperforms the best SAC baselines by approximately \textbf{50}x early in the training process (at 100k iteractions), by \textbf{1.8}x at 300k interactions, and \textbf{1.3}x at 500k interactions.
%
This indicates that AltNet makes more effective use of each environment interaction and achieves higher sample efficiency than SAC at any replay ratio.

\begin{table}[ht!!!]
    \centering
    \caption{Fixed-budget returns at 100k, 300k, and 500k environment steps. AltNet achieves higher returns at every interaction budget, indicating greater sample efficiency compared to baseline SAC at various replay ratios.}
    \label{tab:fixed_budget}
    \begin{tabular}{lccc}
        \toprule
        \textbf{Method} & \textbf{100k Steps} & \textbf{300k Steps} & \textbf{500k Steps} \\
        \midrule
        AltNet (RR = 4) & \textbf{43.7} & \textbf{312.8} & \textbf{385.7} \\
        SAC (RR = 1)              & 0.14 & 106.8 & 190.7 \\
        SAC (RR = 4)              & 0.84 & 171.2 & 254.6 \\
        SAC (RR = 8)              & 0.56 & 79.9  & 294.3 \\
        SAC (RR = 32)             & 0.12 & 120.9 & 161.9 \\
        \bottomrule
    \end{tabular}
\end{table}

\subsection{What Accounts for AltNet's Success?}
\label{sec:abl_buffer}

Having demonstrated AltNet’s advantages, we now investigate the mechanisms behind its success. We first rule out model capacity as a primary factor and then isolate the contributions of alternating networks, resets, and replay-buffer preservation.\\\extraParagraphSpacing{}

% IN this section, we ask questions, to furtehr validate our original hypothesis that. 

\begin{figure}[ht!!!]
    \centering
    \includegraphics[width=1\linewidth]{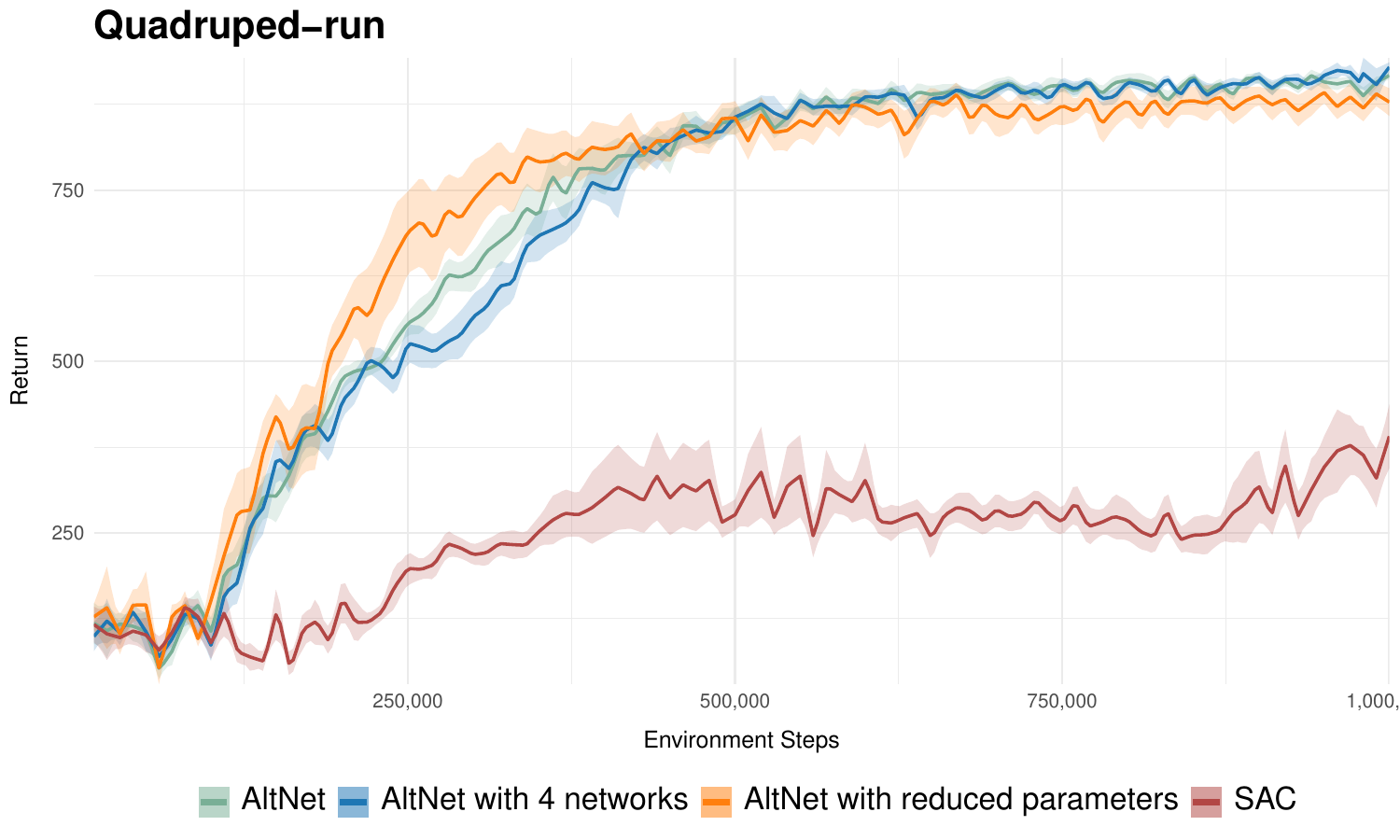}
    \caption{We compare standard AltNet (green), a reduced-parameter version matching SAC’s parameter count (orange), and a scaled variant with four alternating networks (blue), against the SAC baseline (red). Curves report mean episodic return over 5 seeds, with shaded regions indicating $\pm$1 standard error. Despite reduced capacity or increased number of networks, AltNet achieves similar performance, ruling out model size and network count as drivers of its gains.}
    \label{fig:vari}
    % %\vspace{-3ex}
\end{figure}

% \textbf{Increase number of training parameters. } 
\noindent \textbf{RQ1. Does AltNet's advantage arise from increased model capacity?}
A natural hypothesis is that AltNet’s advantage arises simply from the increased number of trainable parameters, since it uses two networks. To test this, we reduced the size of AltNet's networks so that the total number of trainable parameters matches that of a single-network baseline (SAC).
AltNet's performance remained largely unchanged (\autoref{fig:vari}, orange curve).\\\extraParagraphSpacing{}

\noindent \textbf{RQ2. Does increasing the number of networks in AltNet provide additional benefits?}
While AltNet's alternating-network setup is primarily designed to anchor performance during resets, it may also encourage exploration by maintaining two policies that learn in parallel and take turns acting in the environment. This raises the question: can increasing the number of networks beyond two further improve performance? To test this, we modify AltNet to use four networks and adjust the reset frequency accordingly, as specified in
Eq.~\eqref{eq:resetfreq}. As shown in \autoref{fig:vari} (blue curve), scaling beyond two networks does not yield additional gains. We therefore conclude that policy diversity is not a primary contributor to AltNet’s gains.\extraParagraphSpacing{}

\begin{figure}[hb!!]
    \centering
    \includegraphics[width=1\linewidth]{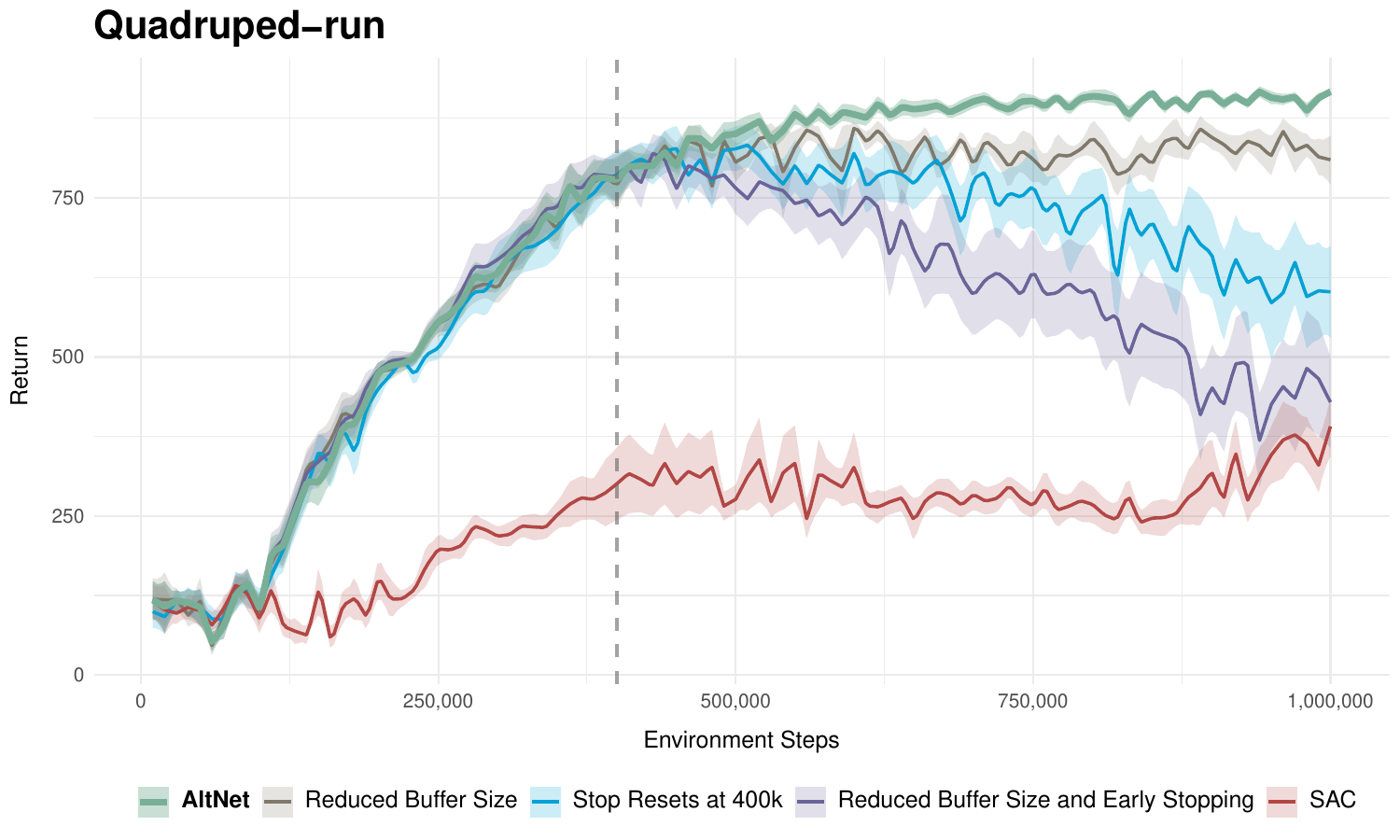}
    % {rka.pdf}
    \caption{
    We compare standard AltNet (green), AltNet with the replay buffer reduced to 400k transitions (gray), AltNet with resets halted after 400k steps (blue), AltNet with both interventions combined (red), and the SAC baseline (purple). Curves report mean episodic return over 10 seeds, with shaded regions indicating $\pm$1 standard error. 
    Both preserving the full replay buffer and maintaining resets are essential for sustaining AltNet’s stability and performance.}
    \label{fig:buffer-reset-400k}
    %\vspace{-3ex}
\end{figure}

The above analyses rule out model size and the number of networks as the primary drivers of AltNet’s gains, so we shift our attention to its defining mechanisms. Specifically, AltNet's performance when instantiated with SAC relies on two interacting processes: (i) preserving and reusing experience across resets via a shared replay buffer, and (ii) periodic alternating resets that restore network plasticity. To examine the role of each component, we perform experiments that disrupt them individually and in combination (\autoref{fig:buffer-reset-400k}).\\\extraParagraphSpacing{}

% \textbf{Impact of Reducing Replay Buffer Capacity. } 
\noindent \textbf{RQ3. Is the preservation of the replay buffer critical to AltNet’s stability?
}The replay buffer plays a critical role in AltNet by providing information continuity during resets. 
% If both the network and buffer were reset simultaneously, training would effectively restart from scratch, eliminating the benefits of alternating resets. 
In the experiments described so far, we have retained every agent interaction in the replay buffer. However, most off-policy algorithms do not preserve the full replay buffer indefinitely. To test the impact of buffer preservation on AltNet's performance, we reduced the replay buffer capacity from the default 1M transitions to 400k, replacing old samples in a first-in, first-out manner. As shown in \autoref{fig:buffer-reset-400k} (gray curve), reducing replay buffer capacity led to a noticeable decline in performance compared to runs with the full buffer. This suggests that preserving the full replay buffer is critical for stabilizing resets and maintaining AltNet’s advantage.\\\extraParagraphSpacing{}

% \textbf{Impact of Halting Alternating Resets. } 
\noindent \textbf{RQ4. Are alternating resets necessary once performance plateaus?
}AltNet provides substantial performance gains during the early stages of training, after which progress slows as performance approaches the task’s achievable optimum. For example, in \texttt{Quadruped-run}, most improvements occur within the first 500k steps, after which performance stabilizes (see \autoref{fig:buffer-reset-400k}). 
This raises a natural question: once learning has plateaued, can a single network---without alternating resets---maintain both plasticity and stability, continuing to perform well over time? To test this, we disabled resets after 400k steps while preserving the full replay buffer. As shown in \autoref{fig:buffer-reset-400k} (blue curve), performance declined once resets were discontinued. 
This indicates that resets remain essential even after apparent convergence, as stopping them may lead to plasticity loss and/or eventual decline in performance. 
%
%This indicates that resets remain essential even after apparent convergence, as stopping them leads to eventual declines in performance during learning, which are apparently resolved by AltNet's resetting. 
% the network can no longer reinforce or adapt its behavior effectively. 
Alternating network resets, by contrast, preserve both plasticity and stability, allowing AltNet to maintain high performance over time.\\\extraParagraphSpacing{}

\noindent \textbf{RQ5. What happens when both mechanisms (buffer preservation and resets) are disrupted simultaneously?
}Finally, we examine the combined effect of disrupting both mechanisms. If resets and buffer preservation are essential for plasticity and stability, then removing both should compound performance degradation. As expected, stopping resets while simultaneously reducing replay buffer size produced the lowest returns, aside from the SAC baseline (\autoref{fig:buffer-reset-400k}, purple curve).\extraParagraphSpacing{}

To verify that these findings are robust to training hyperparameters and reset timing, we repeated the experiment with a different learning rate and a different cutoff timestep for stopping resets and discarding experiences, and observed the same qualitative pattern (see ~\autoref{sec:appendix_additional}). These findings demonstrate that AltNet’s gains are not driven by increased model capacity; it maintains its advantages even when the total number of parameters matches that of the SAC baseline. In addition, increasing the number of networks does not yield any further gains. Rather, AltNet's improvements arise from the interplay between alternating resets and replay buffer preservation. Resets restore plasticity by mitigating failure modes such as persistently inactive neurons, rank collapse, or inflated weight norms (see \autoref{sec:appendix_plasticity}); they reinitialize the network into a more plastic and learnable state, allowing AltNet to better exploit the increasingly representative replay buffer as training progresses. Resets also allow diverse policies to sequentially interact with the environment, which may help with exploration. Alternating networks and preserving the replay buffer maintain learning stability and continuity across resets. We find that disabling either of these components negatively impacts AltNet's performance when paired with SAC. Next, we examine whether AltNet’s benefits extend to on-policy methods, which lack a replay buffer.\extraParagraphSpacing{}

\subsection{AltNet in On-Policy Settings}
\label{sec:ppo}

On-policy reinforcement learning is a critical test of a method’s generality and robustness. In on-policy RL, agents collect trajectories by following their current policy, and updates are performed only on newly collected samples. Agents do not have access to a replay buffer to use past data.\extraParagraphSpacing{}

Existing full network reset methods, such as Standard Resets \citep{nikishin2022primacy} and RDE \citep{kim2023sample}, have been developed and evaluated primarily in off-policy settings. \citet{juliani2024study} showed that plasticity-preserving methods that are effective in off-policy settings often fail to provide benefits when paired with on-policy algorithms, where optimization dynamics differ substantially. Therefore, we now investigate whether AltNet’s benefits extend to on-policy settings, which lack a replay buffer.\extraParagraphSpacing{}

\begin{figure}[ht!]
    \centering
    \includegraphics[width=1\linewidth]{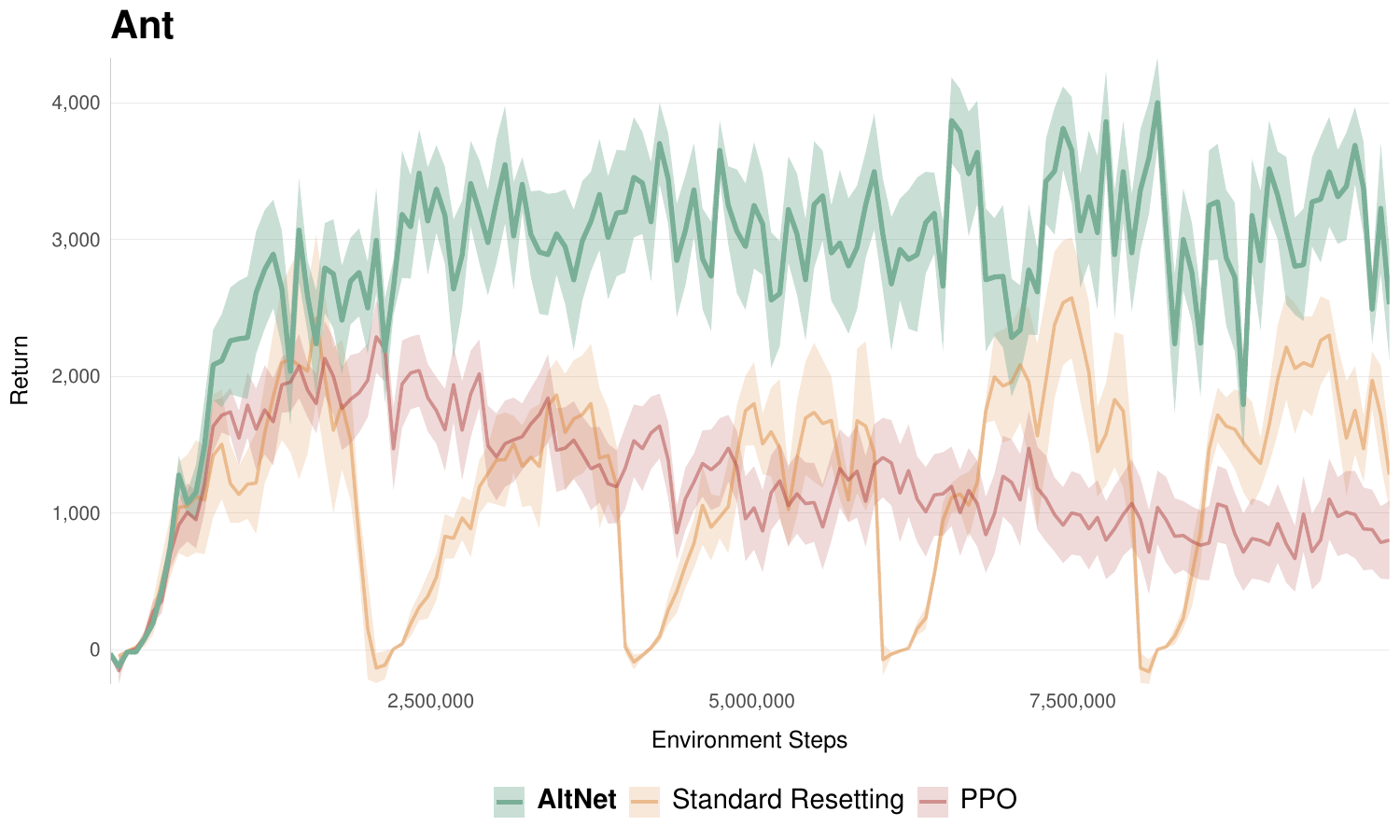}
    \caption{Performance of PPO (red curve), PPO augmented with Standard Resets (orange curve), and AltNet (green curve) in the MuJoCo \texttt{Ant} environment. Curves report mean episodic return over 10 seeds, except for Standard Resets, which was averaged over 5 seeds; shaded regions denote $\pm$1 standard error. AltNet provides consistent gains, demonstrating its efficacy in on-policy settings.}
    \label{fig:ppo_fig}
\end{figure}

We evaluate AltNet in MuJoCo’s Ant environment for 10M steps, a long-horizon setting where PPO suffers severe plasticity loss; in particular, after PPO reaches moderate performance, expected return declines since the agent is incapable of sustaining learning (\autoref{fig:ppo_fig}, red curve). In contrast, AltNet achieves nearly twice the performance of PPO and maintains it indefinitely (\autoref{fig:ppo_fig}, green curve). AltNet’s benefits persist due to an indirect form of knowledge transfer between its networks. While one network interacts with the environment, the other (recently reset) network learns in parallel from the same trajectories, enabling it to recover useful representations. Parallel learning anchors performance across resets and allows AltNet to maintain stability and plasticity in the on-policy regime.\extraParagraphSpacing{}

We further compare AltNet with Standard Resets \citep{nikishin2022primacy}, which, without a replay buffer, suffer post-reset performance collapse (\autoref{fig:ppo_fig}, orange curve). Note that RDE \citep{kim2023sample} is not included in these experiments as it was not designed to tackle on-policy settings. In particular, it relies on a replay buffer and ranks actions according to their Q-values, estimated by off-policy critic. PPO, by contrast, relies on a critic that estimates state values rather than action values. AltNet thus emerges as the only full network reset method capable of restoring plasticity and achieving consistent performance gains across both off-policy and on-policy settings.\extraParagraphSpacing{}

\section{Discussion}
\label{sec:ideal_agent}
While our work focuses on mitigating plasticity loss, it is important to situate plasticity within the broader set of attributes of effective reinforcement learning agents. 
%
%Ideally, RL agents should remain plastic (retaining their ability to update predictions over time), adapt rapidly to distribution shift, make full use of past data, and achieve high performance with limited interactions, while also ensuring performance stability. 
Ideally, RL agents should remain plastic (retaining their ability to update predictions over time) and adapt rapidly to distribution shift. They should also make full use of past data and achieve high performance with limited interactions, while maintaining stable performance.
%
%Plasticity underpins these goals: without the capacity to change, rapid adaptation, data efficiency are negatively impacted. 
Plasticity underpins these goals: without the capacity to change, rapid adaptation and data efficiency are compromised.
%
%Viewed through this lens, AltNet addresses more than plasticity loss. It also enables rapid adaptation, sustained learning from replay buffers, and efficient use of data. 
%
Viewed through this lens, AltNet addresses more than plasticity loss. It also enables rapid adaptation and efficient reuse of data, supporting sustained learning.
%
%Furthermore, by anchoring performance through a twin network, AltNet learns in a stable manner and without performance degradation. 
Furthermore, by anchoring performance with a twin network, AltNet maintains stable learning dynamics and avoids performance degradation.
%
%Together, these capabilities are foundational aspects for the broader qualities that characterize an ideal reinforcement learning agent.
%
Together, these capabilities are foundational for reinforcement learning agents that must continuously adapt over time while remaining stable and data-efficient.
\extraParagraphSpacing{}

\section{Limitations and Future Work}
\label{app:limit}
%Although AltNet demonstrates strong empirical gains and stability across a range of continuous control tasks, our investigation can be extended. We focused our experiments on challenging continuous-action problems from the DeepMind Control Suite, which are representative benchmarks used to evaluate state-of-the-art RL algorithms. That said, extending the evaluation to more diverse environments could provide further insights. Second, AltNet relies on the choice of reset frequency, which currently follows a pre-defined schedule. Developing adaptive scheduling mechanisms to automatically determine the optimal reset frequency, as a function of the environment and replay ratio, is relevant future work.
Although AltNet demonstrates strong empirical gains and stability across a range of continuous-control tasks, there are several directions for further study. Our experiments focus on challenging continuous-action problems from the DeepMind Control Suite, a widely used benchmark for modern reinforcement learning. Extending the evaluation to more diverse environments is likely to lead to further insights. 
In addition, AltNet relies on a choice of predetermined reset frequency. An important direction for future work is to develop adaptive scheduling mechanisms that select reset frequencies automatically based on the environment and replay ratio.
\extraParagraphSpacing{}

% \textcolor{red}{To support reproducibility and further research, we will release the implementation and training scripts upon acceptance.}\\
% Code is available at \url{https://github.com/mansi99000/AltNet}.

%%%%%%%%%%%%%%%%%%%%%%%%%%%%%%%%%%%%%%%%%%%%%%%%%%%%%%%%%%%%%%%%%%%%%%%%

%%% The acknowledgments section is defined using the "acks" environment
%%% (rather than an unnumbered section). The use of this environment 
%%% ensures the proper identification of the section in the article 
%%% metadata as well as the consistent spelling of the heading.

% \begin{acks}
% If you wish to include any acknowledgments in your paper (e.g., to 
% people or funding agencies), please do so using the `\texttt{acks}' 
% environment. Note that the text of your acknowledgments will be omitted
% if you compile your document with the `\texttt{anonymous}' option.
% \end{acks}

%%%%%%%%%%%%%%%%%%%%%%%%%%%%%%%%%%%%%%%%%%%%%%%%%%%%%%%%%%%%%%%%%%%%%%%%

%%% The next two lines define, first, the bibliography style to be 
%%% applied, and, second, the bibliography file to be used.

\bibliographystyle{ACM-Reference-Format} 
\bibliography{sample}

%%%%%%%%%%%%%%%%%%%%%%%%%%%%%%%%%%%%%%%%%%%%%%%%%%%%%%%%%%%%%%%%%%%%%%%%
\clearpage
\newpage
% \end{document}

\appendix

\section*{\LARGE{{Appendix}}}
\vspace{0.3cm}

\section{Hyperparameters Used}
The hyperparameters for AltNet and the baseline methods follow standard continuous-control reinforcement learning practice, with modifications only where required to support resets.
For reproducibility, we report hyperparameter values used when evaluating algorithms in the DMC test suite.\extraParagraphSpacing{}
%The hyperparameters for all algorithms follow standard practice in continuous-control reinforcement learning. 
%
Our PPO implementation is based on the \texttt{CleanRL} library~\cite{huang2022cleanrl} and incorporates AltNet’s alternating reset mechanism.
%Our PPO implementation is based on the \texttt{CleanRL} library~\cite{huang2022cleanrl} and extended with our alternating reset mechanism (AltNet). 
Soft Actor–Critic (SAC) and its reset-based variants (Standard Resets, RDE, and AltNet) were implemented using the \texttt{Stable-Baselines3} framework~\cite{raffin2021stable} following the design decisions introduced in the RDE paper~\cite{kim2023sample}. % on top of which we build our extensions.
The only additional modifications we implemented were those required by AltNet, such as resets and network alternation.
%Only modifications related to resets and network alternation (required by AltNet) were introduced.%; all other settings follow default implementations.
%\vspace{-2mm}

\begin{table}[htbp!]
\caption{Hyperparameters for PPO.}
\label{tab:hyperparams_ppoaltnet}
\begin{center}
\begin{tabular}{ll}
\textbf{Hyperparameters} & \textbf{Value} \\
\hline \\
Environment & \texttt{Ant-v4} (MuJoCo) \\
Total timesteps & $1 \times 10^7$ \\
Seed & 1 \\
Number of environments & 1 \\
\hline \\
Algorithm & PPO (CleanRL) \\
Number of networks (AltNet) & 2 (active / passive) \\
Reset frequency (environment steps) & $2 \times 10^5$ \\
Optimizer & Adam \\
Optimizer: learning rate & 0.0003 \\
Optimizer: $\epsilon$ & $1 \times 10^{-5}$ \\
Learning rate annealing & False \\
Discount factor ($\gamma$) & 0.99 \\
GAE $\lambda$ & 0.95 \\
\hline \\
Rollout length (per env) & 2048 steps \\
Number of minibatches & 32 \\
Number of update epochs & 10 \\
Batch size & 2048 \\
Minibatch size & 64 \\
Advantage normalization & True \\
\hline \\
Clip coefficient ($\epsilon$) & 0.2 \\
Value loss coefficient ($c_v$) & 0.5 \\
Entropy coefficient ($c_e$) & 0.0 \\
Max gradient norm & 0.5 \\
Target KL divergence & None \\
\hline \\
Network architecture & MLP (64–64) \\
Trainable log standard deviation & Initialized to 0 \\
\end{tabular}
\end{center}
\end{table}

\begin{table}[t]
\caption{Hyperparameters for SAC and variants.}
\label{tab:hyperparams1}
\begin{center}
\begin{tabular}{ll}
\textbf{Hyperparameters} & \textbf{Value} \\
\hline \\
\# of network (AltNet and RDE) & 2 \\
\# of network (Baseline and Standard Reset) & 1 \\
Training steps & $1 \times 10^6$ \\
Discount factor & 0.99 \\
Warm up period & 5000 \\
Minibatch size & 1024 \\
Optimizer & Adam \\
Optimizer : learning rate & 0.0003 \\
Networks : activation & ReLU \\
Networks : n. hidden layers & 2 \\
Networks : neurons per layer & 1024 \\
Initial Temperature & 1 \\
Replay Buffer Size & $1 \times 10^6$ \\
Updates per step (Replay Ratio) & (1, 4) \\
Target network update period & 1 \\
$\tau$ (Polyak update) & 0.005 \\
\hline \\
Reset Frequency (gradient steps) for all & $2 \times 10^5$ \\
$\beta$ (action select coefficient) for RDE & 50 \\
\end{tabular}
\end{center}
\end{table}

\section{AUC Calculation}
\label{sec:auc}
We use the \emph{area under the learning curve (AUC)} to quantify overall learning efficiency. This measure reflects both how quickly and how effectively an agent learns, combining stability and final performance into a single scalar. Higher AUC indicates that agents not only achieve high performance but also maintain it consistently throughout training. %\extraParagraphSpacing{}
To summarize the overall learning trajectory of each agent, we compute the AUC of average return over time.
For a learning curve defined by episodic returns $R_t$ recorded at discrete training steps $t \in [t_0, t_T]$, the AUC is given by
% \[
% \text{AUC}_{\text{raw}} = \sum_{i=1}^{T-1} \frac{R_{t_i} + R_{t_{i+1}}}{2} \, (t_{i+1} - t_i),
% \]
\begin{equation}
\text{AUC}_{\text{raw}} = \sum_{i=1}^{T-1} \frac{R_{t_i} + R_{t_{i+1}}}{2} (t_{i+1}-t_i),
\end{equation}
which approximates the integral of performance over time using the trapezoidal rule. For readability and comparability, we report the \emph{normalized} AUC, obtained by dividing $\text{AUC}_{\text{raw}}$ by the total number of training steps ($t_T - t_0 = 1\text{M}$). These normalized values represent the average return over the training horizon rather than the raw cumulative sum of rewards. Note that normalization does not affect relative rankings since all runs share the same training horizon; instead, it yields interpretable measurements suitable for cross-environment comparison, as reported in \autoref{tab:auc_all}.

\section{STATISTICAL ANALYSIS OF NORMALIZED AUC}\label{app:bootstrap}

To quantify uncertainty in the normalized AUC values reported in Table~1, we compute 95\% confidence intervals using the stratified bootstrap method. For each condition (defined by a combination of environment and replay ratio), we treat the per-seed normalized AUC values as the bootstrap samples. 
Specifically, for each condition and method, we resample the $n$ per-seed AUC values with replacement to form a bootstrap sample of size $n$, compute its mean, and repeat this procedure $B = 10{,}000$ times. The 95\% confidence interval is defined using the 2.5th and 97.5th percentiles of the resulting bootstrap distribution of means. 
%
%This procedure is stratified in the sense that resampling is performed independently within each \{environment, replay ratio, method\} stratum, preserving the experimental structure across conditions. The normalized AUC for each seed is computed by interpolating the learning curve onto 100 evenly-spaced evaluation points spanning the full training range and taking the mean return across those points, yielding an interpretable ``average return over training'' metric consistent with the normalization described in \autoref{sec:auc}.\extraParagraphSpacing{}
%
This procedure is stratified by resampling independently within each \{environment, replay ratio, method\} stratum, which preserves the experimental structure across conditions. For each seed, we compute the normalized AUC by interpolating the learning curve onto 100 evenly spaced evaluation points spanning the full training horizon and then averaging the returns across these points. This yields an interpretable ``average return over training'' metric consistent with the normalization described in~\autoref{sec:auc}.\extraParagraphSpacing{}
We assess statistical significance using non-overlapping confidence intervals: if the lower bound of one method's CI exceeds the upper bound of another, we conclude that the difference is statistically significant at the 95\% level.~\autoref{tab:bootstrap_ci} reports the full results.\\\extraParagraphSpacing{}

\begin{table}[hb!!]
\centering
\caption{Normalized AUC with 95\% stratified bootstrap confidence intervals. The best method per condition is highlighted in \textbf{bold}. All methods use 10 seeds.}
\label{tab:bootstrap_ci}
\small
\begin{tabular}{llcc}
\hline
\textbf{Environment} & \textbf{Method} & \textbf{Mean} & \textbf{95\% CI} \\
\hline
\multirow{4}{*}{Cheetah (RR=1)}
 & \textbf{AltNet} & \textbf{656} & [631, 681] \\
 & SAC             & 614 & [592, 637] \\
 & SR              & 527 & [497, 554] \\
 & RDE             & 595 & [567, 624] \\
\hline
\multirow{4}{*}{Cheetah (RR=4)}
 & \textbf{AltNet} & \textbf{719} & [694, 741] \\
 & SAC             & 534 & [437, 621] \\
 & SR              & 618 & [599, 638] \\
 & RDE             & 597 & [580, 614] \\
\hline
\multirow{4}{*}{Hopper (RR=1)}
 & \textbf{AltNet} & \textbf{216} & [179, 256] \\
 & SAC             & 157 & [126, 189] \\
 & SR              & 154 & [140, 167] \\
 & RDE  & 188 & [171, 207] \\
\hline
\multirow{4}{*}{Hopper (RR=4)}
 & \textbf{AltNet} & \textbf{313} & [262, 368] \\
 & SAC             & 205 & [142, 258] \\
 & SR              & 248 & [204, 296] \\
 & RDE             & 277 & [237, 321] \\
\hline
\multirow{4}{*}{Quadruped (RR=1)}
 & \textbf{AltNet} & \textbf{618} & [596, 639] \\
 & SAC             & 377 & [297, 461] \\
 & SR              & 568 & [540, 595] \\
 & RDE             & 608 & [592, 623] \\
\hline
\multirow{4}{*}{Quadruped (RR=4)}
 & AltNet          & 702 & [686, 718] \\
 & SAC             & 241 & [214, 269] \\
 & SR              & 686 & [662, 710] \\
 & \textbf{RDE}    & \textbf{715} & [702, 727] \\
\hline
\multirow{4}{*}{Walker (RR=1)}
 & \textbf{AltNet} & \textbf{646} & [632, 659] \\
 & SAC   & 568 & [527, 606] \\
 & SR              & 615 & [602, 628] \\
 & RDE  & 457 & [263, 646] \\
\hline
\multirow{4}{*}{Walker (RR=4)}
 & \textbf{AltNet} & \textbf{725} & [715, 736] \\
 & SAC             & 652 & [636, 665] \\
 & SR              & 722 & [713, 730] \\
 & RDE             & 721 & [715, 726] \\
\hline
\end{tabular}

\vspace{4pt}
% {\footnotesize $^\dagger$9 seeds. \quad $^*$9 seeds. \quad $^\ddagger$6 seeds.}
\end{table}

% \\\extraParagraphSpacing{}
\noindent \textbf{Key findings.}
\autoref{tab:bootstrap_ci} supports several conclusions about AltNet's performance relative to baselines.\extraParagraphSpacing{}
First, \emph{AltNet is never significantly outperformed.} Across all 8 conditions (i.e., combinations of environments and replay ratios), no competing method's CI lower bound exceeds AltNet's CI upper bound. AltNet achieves the highest, or a statistically tied-for-highest, normalized AUC in every condition.\extraParagraphSpacing{}

Second, \emph{AltNet significantly outperforms SAC in 6 of 8 conditions.} In Cheetah (RR=4), Hopper (RR=4), Walker (RR=1 and RR=4), and Quadruped (RR=1 and RR=4), AltNet's CI lower bound exceeds SAC's CI upper bound---the intervals do not overlap. In the two exceptions where CIs overlap (Cheetah with RR=1 and Hopper with RR=1), AltNet's mean AUC is higher. These observations confirm that the aggregate 38\% improvement reported in Table~1 reflects genuine performance differences rather than seed-level noise.\extraParagraphSpacing{}

Third, \emph{AltNet significantly outperforms Standard Resets in 5 of 8 conditions.} AltNet's performance CI is strictly above SR's in Cheetah (RR=1 and RR=4), Hopper (RR=1), Walker (RR=1), and Quadruped (RR=1). In the remaining 3 conditions, AltNet's mean AUC is higher, though 
%normalized AUC values' 
CIs overlap partially. 
Importantly, in cases where performance CIs overlap, it is possible to observe that Standard Resets often suffers recurring performance collapses after each reset (see Figures~2 and~3, orange curves), whereas AltNet maintains stable performance throughout training. 
%This analysis does not capture a critical qualitative difference: Standard Resets suffer recurring performance collapses after each reset (see Figures~2 and~3, orange curves), whereas AltNet maintains stable performance throughout training. In conditions where AUC values are comparable, this stability advantage is masked by the aggregate metric---SR's post-reset drops are offset by recovery periods, yielding similar area under the curve. 
%
%For safety-critical or deployment-sensitive settings where transient performance collapses are unacceptable, AltNet provides a preferable solution even when the aggregate AUC is similar.
%
\extraParagraphSpacing{}

Fourth, \emph{RDE is AltNet's closest competitor in aggregate AUC.} AltNet significantly outperforms RDE in 2 of 8 conditions (namely, the Cheetah settings with RR=1 and RR=4). In both, AltNet's CI is strictly above RDE's. In the remaining 6 conditions, CIs overlap. Notably, in Quadruped (RR=4), RDE achieves a slightly higher sample mean AUC (715 vs.\ 702), consistent with the observation in~\autoref{sec:results} that RDE is competitive in this environment. Importantly, though, in this latter case, the CIs overlap substantially. Furthermore, AltNet achieves this competitive performance with a substantially simpler architecture. 
RDE maintains an ensemble of networks with a Q-value-weighted action selection mechanism and an additional hyperparameter ($\beta$) that governs gating behavior, whereas AltNet uses only two networks with a deterministic alternation schedule and no additional hyperparameters beyond the reset frequency. 
Furthermore, RDE allows recently reset networks to act in the environment, leading to residual post-reset performance drops (see Figure~2, blue curves), whereas AltNet significantly reduces these drops by construction. Thus, in cases where the two methods yield comparable AUC, AltNet does so with greater simplicity, fewer hyperparameters, and superior stability.\extraParagraphSpacing{}

Finally, \emph{Hopper exhibits the widest confidence intervals.} AltNet's CIs  span $[179, 256]$ (at RR=1) and $[262, 368]$ (at RR=4). Such wide CIs result from the high inter-seed variance characteristic of the hopping domain. Despite this spread, AltNet significantly outperforms both SAC and SR at RR=4 and SR at RR=1. All other pairwise comparisons in Hopper involve overlapping CIs, reflecting the inherent stochasticity of this environment rather than a limitation specific to AltNet.\\\extraParagraphSpacing{}

\noindent \textbf{Summary.}
AltNet achieved the highest (or statistically tied-for-highest) normalized AUC in all 8 conditions we investigated. It significantly outperformed SAC in 6 of 8 conditions, Standard Resets in 5 of 8, and RDE in 2 of 8 conditions. No competing method significantly outperformed AltNet in any condition. Moreover, in conditions where AltNet's AUC confidence intervals overlapped with those of SR or RDE, AltNet retained important qualitative advantages: it eliminated the post-reset performance collapses exhibited by both SR and RDE (Figures~2 and~3), and achieved competitive performance with a simpler architecture and fewer hyperparameters than RDE. Together, these results provide statistical support for the conclusions drawn from the point estimates in Table~1 and reinforce that AltNet offers the best combination of performance, stability, and simplicity among the methods evaluated.

\section{Additional Sample Efficiency Results}
\label{appendix:sample_efficiency}

\begin{figure}[h!!]
    \centering
    \includegraphics[width=1\linewidth]{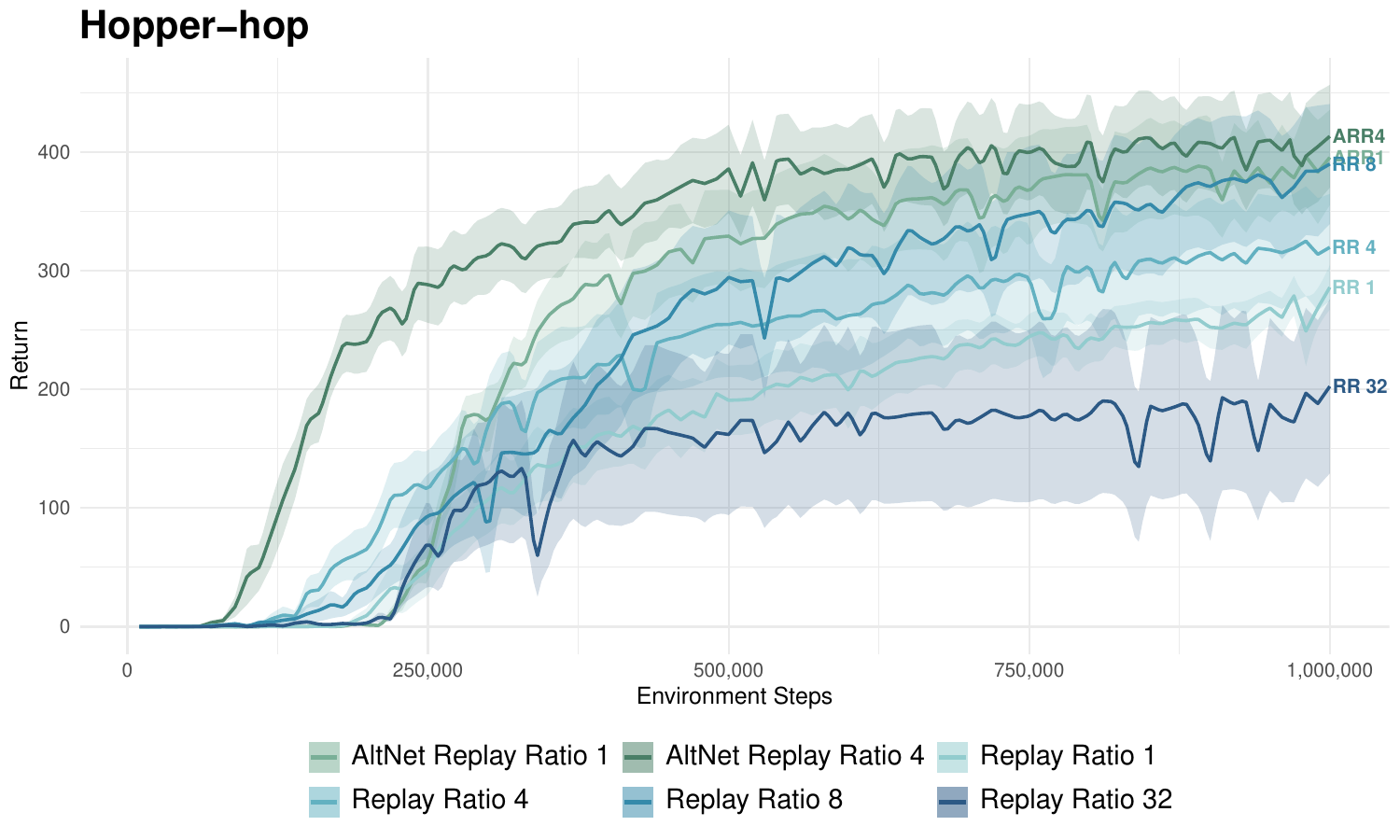}
    \caption{Learning curves of SAC in the \texttt{hopper-hop} environment (DMC) under different replay ratios (RR = 1, 4, 8, 32) and AltNet at RR = 1 and RR = 4. Curves show mean episodic return over 10 seeds, with shaded regions denoting $\pm$1 standard error. For SAC, performance improves as RR increases up to 8, but degrades at RR = 32. AltNet achieves the highest performance (and is the most sample efficient) at RR = 4.}
    \label{fig:all_curves}
    %\vspace{4mm}
\end{figure}

This section provides additional results related to ~\autoref{sec:sample_efficiency}, which examines how AltNet improves performance and sample efficiency compared to SAC. While the main paper (\autoref{sec:sample_efficiency}) focuses on AltNet at $\text{RR}=4$, here we include additional learning curves for AltNet trained at $\text{RR}=1$ to provide a complete view of its behavior across replay ratios. %\extraParagraphSpacing{}
In particular, ~\autoref{fig:all_curves} extends ~\autoref{fig:highRR} from the main text by depicting AltNet’s performance at both $\text{RR}=1$ and $\text{RR}=4$, alongside SAC's performance when trained at $\text{RR}\in\{1,4,8,32\}$. These results confirm the same qualitative patterns observed earlier: as the replay ratio increases, SAC's performance initially improves, but it degrades at very high $\text{RR}$, while AltNet consistently achieves superior performance even at the lowest replay ratio.%\extraParagraphSpacing{}
Together, these additional empirical results reinforce the findings of ~\autoref{sec:sample_efficiency}: AltNet achieves higher performance and greater sample efficiency without requiring aggressive replay schedules or the associated computational overhead.
%\vspace{4mm}

\section{Additional Experiments}
\label{sec:appendix_additional}

This section provides supplementary analyses supporting the results discussed in \autoref{sec:abl_buffer} (``\textit{What Accounts for AltNet's Success?}'').
There, we examined how AltNet’s performance depends on two interacting mechanisms: alternating resets and replay-buffer preservation. To verify that those findings are robust to training hyperparameters and reset timings, we repeat the same ablation experiments under modified conditions.\extraParagraphSpacing{}
Specifically, we vary the reset timing by halting resets at 600k steps instead of 400k to test whether AltNet’s dependence on resets is sensitive to the cutoff point. We also vary the learning rate, reducing it from 0.0003 to 0.0002, to confirm that the observed trends are not optimizer-specific.

\begin{figure}[htbp]
    \centering
    \begin{subfigure}{0.9\columnwidth}
        \centering
        \includegraphics[width=\linewidth]{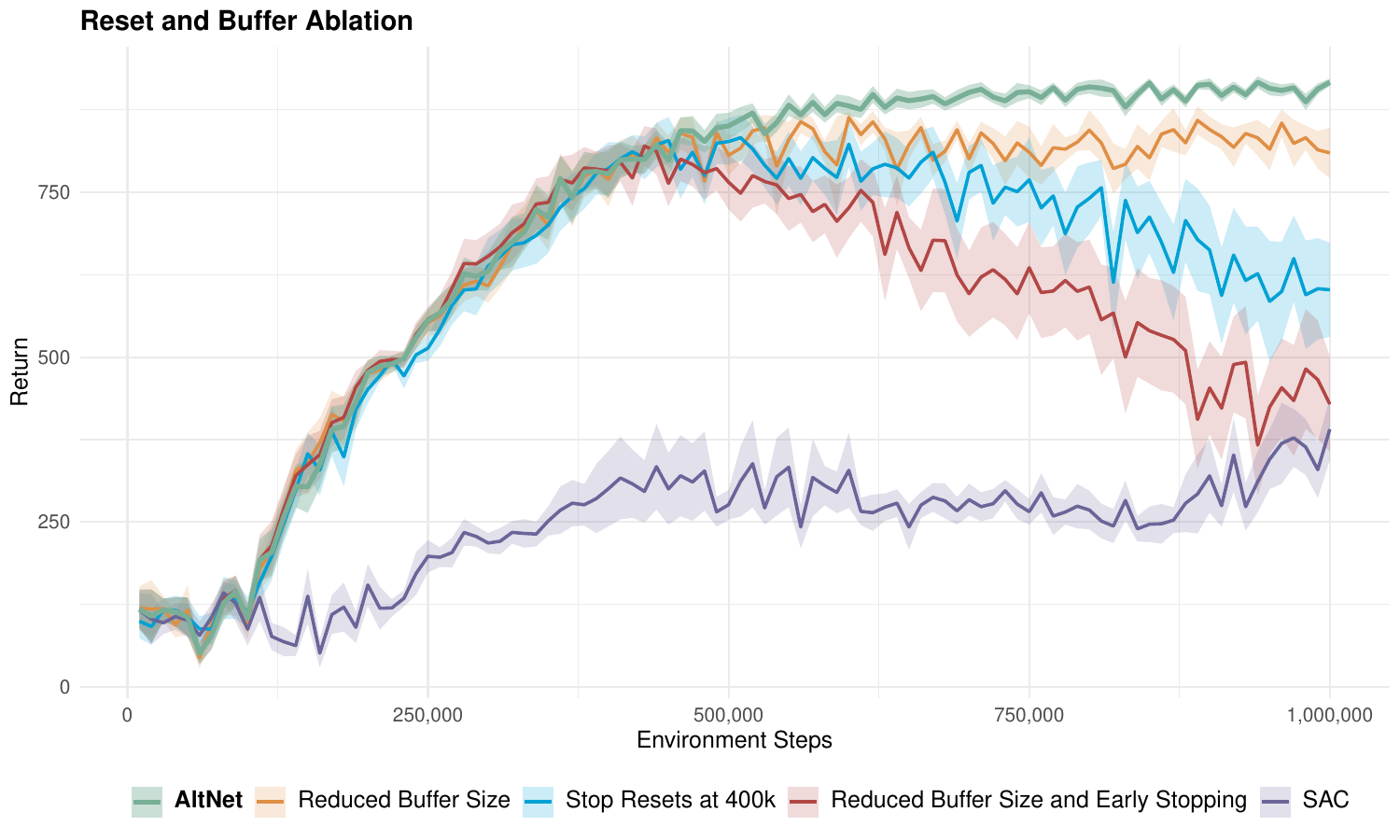}
        \caption{Resets halted after 400k steps. Buffer size reduced to 400k.}
        \label{ffig:buffer-reset-ablation}
    \end{subfigure}
    \hfill
    \begin{subfigure}{0.9\columnwidth}
        \centering
        \includegraphics[width=\linewidth]{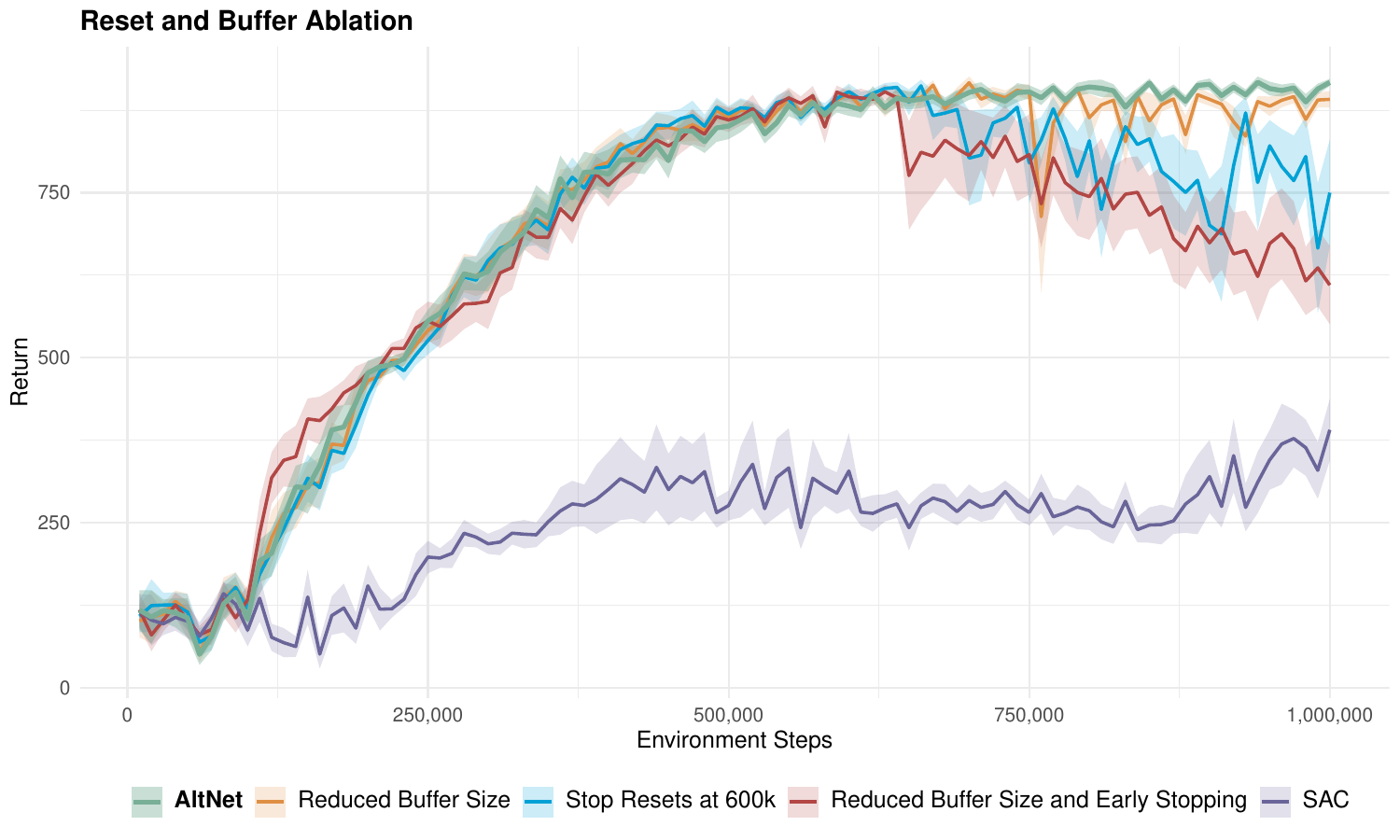}
        \caption{Resets halted after 600k steps. Buffer size reduced to 600k.}
        \label{fig:stop-reset-600k}
    \end{subfigure}
    \caption{Analysis in the \texttt{Quadruped-run} environment (DMC). Curves show mean episodic return over 10 seeds, with shaded regions denoting $\pm$1 standard error. We compare standard AltNet (green), AltNet with reduced buffer size (orange), AltNet with resets halted (blue), AltNet with both interventions combined (red), and SAC (purple). Results demonstrate that both preserving the replay buffer and maintaining resets are essential for AltNet’s stability.}
    \Description{Two subfigures comparing AltNet performance under reduced buffer size and halted resets at 400k and 600k steps, showing stability differences.}
    \label{fig:buffer-reset-ablation}
\end{figure}

\begin{figure}[htbp]
    \centering
    \includegraphics[width=0.8\linewidth]{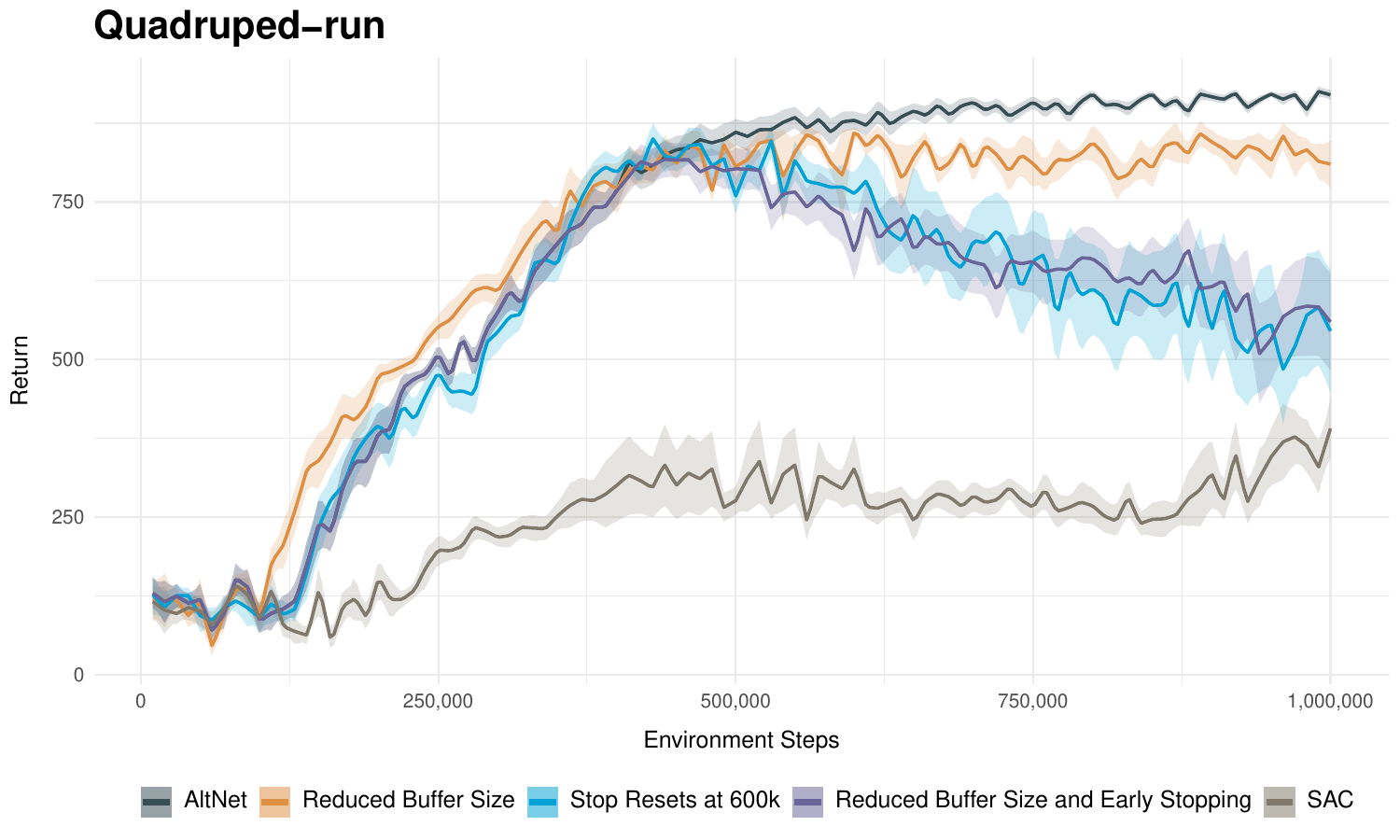}
    \caption{Same replay buffer–reset ablation experiment as ~\autoref{fig:buffer-reset-ablation}(a), now conducted at a lower learning rate (0.0002) to test AltNet’s performance robustness to optimizer settings. We compare standard AltNet (black), AltNet with reduced replay buffer size (orange), AltNet with halted resets (blue), AltNet with both reduced buffer size and halted resets (purple), and the SAC baseline (brown).}
    \label{fig:rep}
    %\vspace{-2mm}
\end{figure}

\autoref{fig:buffer-reset-ablation} compares AltNet, AltNet with reduced-buffer, AltNet with halted-reset, and AltNet with both interventions in the \emph{Quadruped-run} environment for both 400k and 600k step halts. The qualitative empirical findings remain unchanged: performance declines sharply when either resets or buffer preservation is removed, and most severely when both are disabled. Furthermore,~\autoref{fig:rep} depicts results from the same experiment but at a lower learning rate; we again observe consistent behavior. Together, these additional experiments reinforce the conclusions of ~\autoref{sec:abl_buffer}: AltNet’s advantages are not artifacts of specific hyperparameter choices or reset schedules. Network alternation and replay-buffer preservation remain essential for maintaining both plasticity and stability across conditions.\extraParagraphSpacing{}
%\vspace{-6mm}
%\vspace{-2mm}

\section{Plasticity Loss and Resets}
\label{sec:appendix_plasticity}

Plasticity loss refers to a network’s declining ability to learn from new data. A network is said to have lost plasticity when it can no longer optimize its objective as effectively as a freshly initialized counterpart~\citep{lyle2024disentangling}. 
Prior work has shown that this decline is not always evident from return curves alone, and researchers have therefore developed metrics that correlate with plasticity loss, such as increasing weight norms, the emergence of dormant (inactive) units, and reduced representational diversity as measured by stable rank or activation sparsity~\citep{dohare2024loss, lyle2023understanding, sokar2023dormant}. These metrics serve as diagnostic signals for representational collapse, a phenomenon in which the network’s neurons can no longer encode distinct patterns for different inputs and instead produce highly similar or repetitive outputs.\extraParagraphSpacing{}

\begin{figure}[htbp]
    \centering
    \includegraphics[width=0.8\linewidth]{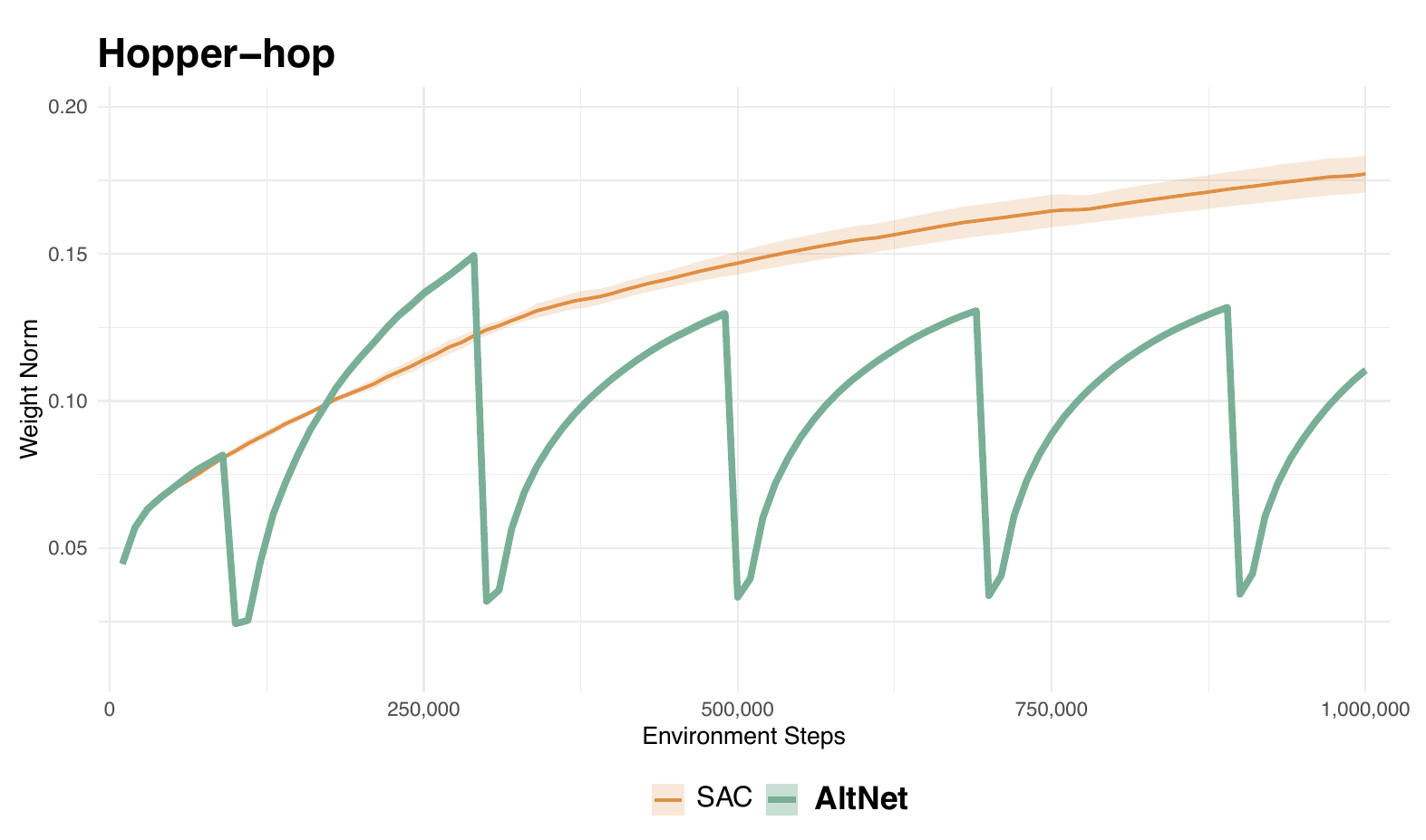}
    \caption{Evolution of the average $\ell_2$ weight norm during training. SAC shows steady growth in weight magnitude over time, reflecting instability and the accumulation of representational drift. AltNet's resets periodically restore weight norms to lower, well-conditioned values, preventing runaway weight growth and preserving learnability.}    
    \label{fig:weight-norm}
    %\vspace{-2mm}
\end{figure}

\begin{figure}[htbp]
    \centering
    \includegraphics[width=0.8\linewidth]{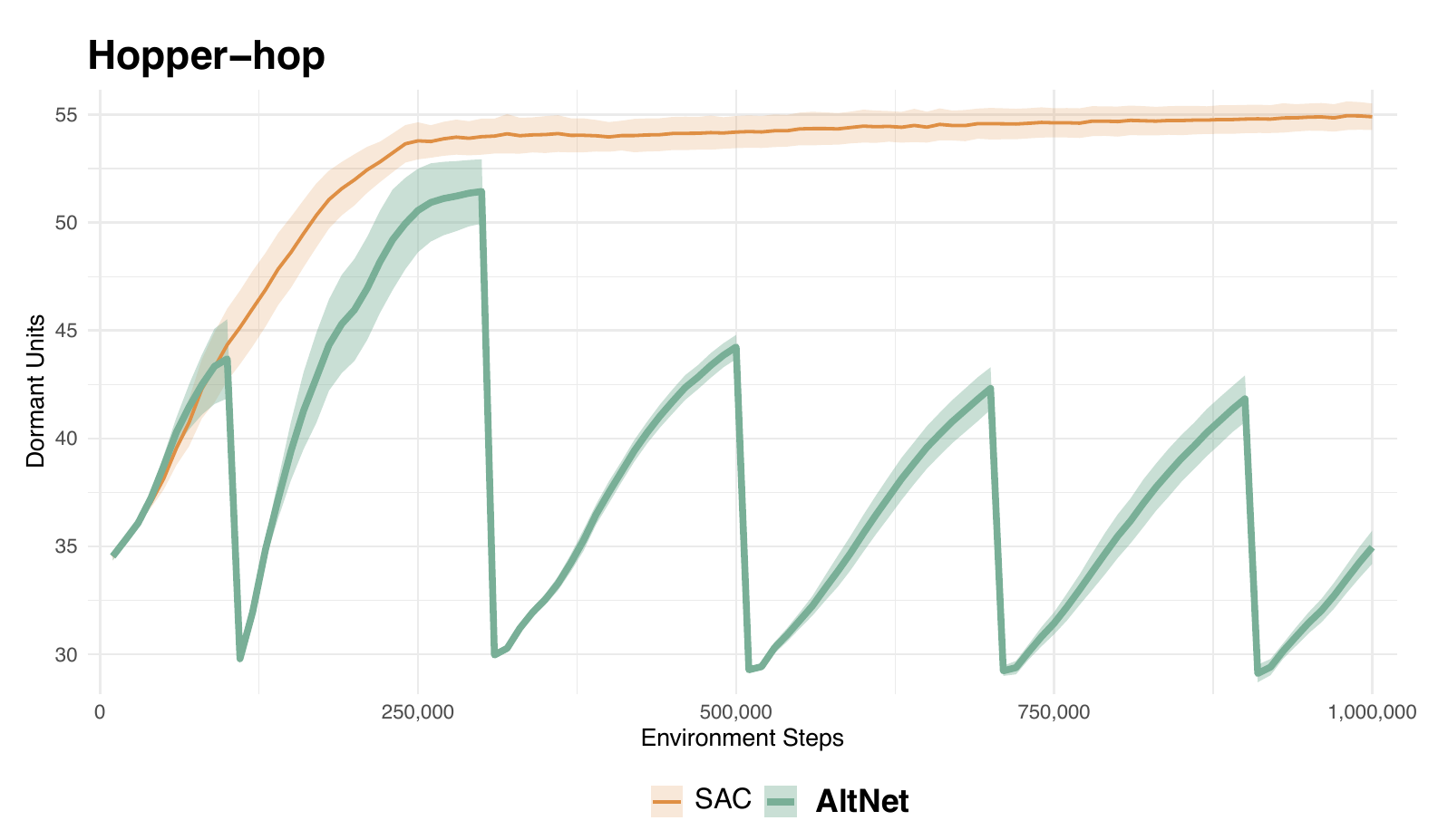}
    \caption{Proportion of dormant (inactive) neurons over training steps. In SAC, the number of inactive neurons increases steadily, indicating loss of network functional capacity. AltNet's resets prevent this accumulation by refreshing neuron activations, thereby preserving functional plasticity throughout training.}
    \label{fig:dormant-units}
    %\vspace{-2mm}
\end{figure}

\begin{figure}[htbp]
    \centering
    \includegraphics[width=0.8\linewidth]{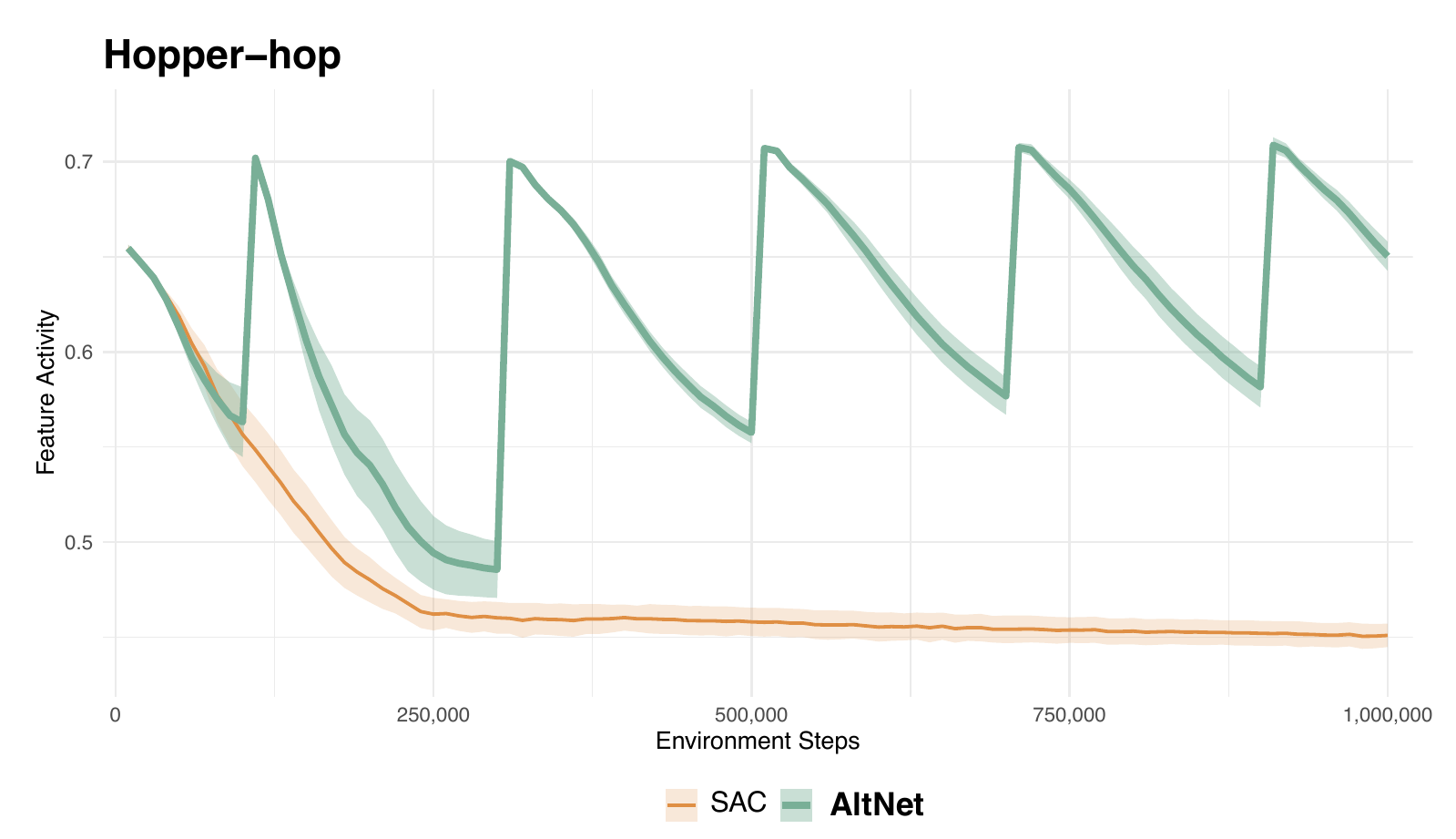}
    \caption{Stable rank during training for SAC and AltNet, which serves as a proxy for representational diversity. In SAC, the stable rank decreases over time, indicating reduced representational diversity. By contrast, AltNet counteracts this trend through periodic reinitialization, maintaining higher diversity throughout training.}
    \label{fig:stable-rank}
    %\vspace{-2mm}
\end{figure}

In our experiments, we tracked these metrics throughout training. Consistent with prior findings, we observed that standard agents, such as those trained with SAC, gradually accumulate internal pathologies, including increasing weight norms (\autoref{fig:weight-norm}), a growing fraction of dormant units (\autoref{fig:dormant-units}), and decreasing representational diversity (\autoref{fig:stable-rank}). 
Together, these trends signal a loss of plasticity, even when external performance initially appears stable. Additionally, we also examined how resets affect these plasticity-correlated metrics. As shown in ~\autoref{fig:weight-norm}--\autoref{fig:dormant-units}, each AltNet full reset restores the network to a state with lower weight norms, fewer dormant units, and higher feature activity. This observation supports the hypothesis that resets reinitialize the network to a well-conditioned, highly plastic state. 


\section{Relationship to Continual Backprop}
\label{app:cbp}

We view Continual Backprop (CBP)~\cite{dohare2024loss} as complementary to AltNet rather than as a direct competitor. The authors of CBP explicitly frame their contribution as complementary to methods addressing our setting---i.e., methods that leverage replay buffers---noting that CBP assumes ``systems that use small or no replay buffers''. The two methods therefore address complementary replay regimes and are not directly comparable. AltNet relies on a persistent replay buffer and is not applicable to CBP's target regime of small-buffer or buffer-free PPO-style systems. Conversely, CBP is not designed to address settings where balancing the interaction between large replay buffers and network resets is paramount to ensure plasticity and fast adaptation.\extraParagraphSpacing{}

\looseness-1
After thoroughly investigating our five main research questions (RQ1--RQ5), we conducted preliminary experiments to evaluate AltNet in on-policy settings where replay buffers cannot be leveraged; i.e., settings that do not satisfy its main assumptions. 
%These exploratory experiments (\autoref{sec:ppo}) were designed to test whether AltNet could potentially be extended to on-policy settings that lack a replay buffer---i.e., settings that do not satisfy its main assumptions. 
%
Our preliminary findings (\autoref{sec:ppo}) highlight promising future directions, including potential comparisons and combinations with CBP.\extraParagraphSpacing{}
%%%%%%%%%%%%%%%%%%%%%%%%%%%%%%%%%%%%%%%%%%%%%%%%%%%%%%%%%%%%%%%%%%%%%%%%
\clearpage
\end{document}